\PassOptionsToPackage{table}{xcolor}
\documentclass{arxiv_singlecolumn}
\microtypesetup{expansion=false}

\usepackage{amsmath,amsthm,amsfonts,bm}
\newtheorem{theorem}{Theorem}
\newtheorem{corollary}[theorem]{Corollary}
\newtheorem*{rankpreservationrestate}{Theorem~\ref*{thm:rank_preservation}}

\def\eqref#1{equation~\ref{#1}}

\def\1{\bm{1}}

\DeclareMathAlphabet{\mathsfit}{\encodingdefault}{\sfdefault}{m}{sl}
\SetMathAlphabet{\mathsfit}{bold}{\encodingdefault}{\sfdefault}{bx}{n}

\newcommand{\KL}{D_{\mathrm{KL}}}

\usepackage{amssymb}
\usepackage{booktabs}
\usepackage{multirow}
\usepackage{threeparttable}
\usepackage{float}
\usepackage{algorithm}
\usepackage{algorithmic}
\usepackage{needspace}
\usepackage{enumitem}
\usepackage{pifont}
\usepackage{bbding}
\usepackage{xcolor}
\usepackage{colortbl}
\definecolor{pink}{HTML}{747199}
\usepackage{graphicx}
\usepackage{wrapfig}
\usepackage{hyperref}
\usepackage{caption}
\usepackage{subcaption}
\usepackage{longtable}
\usepackage{arydshln}
\definecolor{experimentgray}{HTML}{F0F0F0}
\definecolor{experimentours}{HTML}{ECF0FF}
\usepackage{url}
\newlength{\motivationfigureheight}
\definecolor{casequerycolor}{RGB}{68,114,196}
\definecolor{casesatisfycolor}{RGB}{79,173,91}
\definecolor{casejudgecolor}{RGB}{237,125,49}
\newcommand{\casequery}[1]{\textcolor{casequerycolor}{#1}}
\newcommand{\casesatisfy}[1]{\textcolor{casesatisfycolor}{#1}}
\newcommand{\casejudge}[1]{\textcolor{casejudgecolor}{\textbf{#1}}}
\newlength{\caselabelwidth}
\newlength{\casecheckwidth}
\newlength{\casetextwidth}
\newlength{\casespanwidth}

\usepackage[most]{tcolorbox}
\usepackage{needspace}

\tcbuselibrary{breakable}
\tcbuselibrary{skins}
\newtcolorbox[use counter=tcboxcounter,number within=section]{mybox}[3][]{
    left=3pt,
    right=4pt,
    breakable,
    enhanced,
    title=#2 \thetcbcounter: #3,
    #1
}

\newcommand{\arxivMethodTitle}{\textcolor{black}{Rubrics on Trial}}
\newcommand{\githubicon}{\raisebox{-1.5pt}{\includegraphics[height=1.03em]{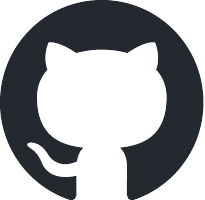}}}
\title{\arxivMethodTitle: Evolving Rubrics from a Single Query via Synthetic Pairwise Evidence}
\author[1]{Haocheng Yang}
\author[2]{Licheng Pan}
\author[3]{Yuan Lu}
\author[4]{Xiaoxi Li}
\author[5]{Zhichao Chen}
\author[5]{Chunyuan Zheng}
\author[6]{Shuai Shao}
\author[5]{Haoxuan Li}
\author[7,8,*]{Hao Wang}
\author[7,8]{Kun Zhang}

\affiliation[1]{National University of Singapore}
\affiliation[2]{Zhejiang University}
\affiliation[3]{Xiaohongshu}
\affiliation[4]{Renmin University of China}
\affiliation[5]{Peking University}
\affiliation[6]{Shanghai Jiaotong University}
\affiliation[7]{MBZUAI}
\affiliation[8]{\mbox{Carnegie Mellon University}}
\contribution[*]{Corresponding author.}
\correspondence{\href{mailto:haocheng_yang@u.nus.edu}{haocheng\_yang@u.nus.edu}, \href{mailto:Ho-ward@outlook.com}{Ho-ward@outlook.com}}
\metadata[\githubicon~Code]{\href{https://github.com/yhc-666/Rubrics-On-Trial.git}{https://github.com/yhc-666/Rubrics-On-Trial.git}}

\abstract{
Rubric evolution offers a promising approach to improving the quality of rubrics generated by large language models (LLMs). Central to this process is rubric comparison, which identifies the better of two rubrics and guides the direction of evolution. However, accurate rubric comparison is difficult, which presents two challenges.
\textbf{\ding{182} It should reflect downstream task performance}, which is essential for assessing rubric utility but often prohibitively expensive to evaluate. \textbf{\ding{183} It should discourage unnecessary criteria}, which increase verification costs and may dilute the influence of essential criteria.
To address these challenges, we introduce \textbf{Rubrics on Trial}, a multi-agent framework that evolves rubrics by comparing synthetic response pairs. To address challenge \ding{182}, the framework compares synthetic responses that satisfy the respective rubrics, providing a proxy for downstream performance without training a separate policy for each rubric. 
To address challenge \ding{183}, it assesses the necessity of a candidate criterion by independently generating high-quality alternative responses that violate it and comparing them with edited versions that satisfy it. A rubric is favored when it improves response quality in both comparisons, and the resulting comparison signal is further incorporated for rubric evolution. Extensive experiments demonstrate that Rubrics on Trial improves the quality of generated rubrics and leads to better downstream task performance.
}

\begin{document}
\maketitle

\section{Introduction}
\label{sec:introduction}

A rubric is a structured set of criteria specifying the requirements an LLM response should meet to satisfy a user query~\citep{hashemi2024llmrubric}. By translating user intent into explicit criteria, rubrics enable fine-grained, verifiable assessment of LLM responses. These properties make rubrics well suited for providing reward signals in reinforcement learning from human feedback (RLHF)~\citep{gunjal2026rubrics,wangunbiased,wang2026causalrm}. Currently,  rubrics have been a critical component for LLM evaluation and  training.

Early efforts to construct rubrics relied on human annotators to formulate evaluation criteria~\citep{arora2025healthbench}. Although effective, it is difficult to extend across diverse user queries and tasks. Subsequent studies used LLMs to generate query-specific rubrics directly, providing a scalable alternative~\citep{cook2024ticking,wei2025rocketeval}. Nevertheless, these approaches may overlook important criteria and do not inherently support iterative refinement of the resulting rubrics~\citep{dhole2026rubricrag}. To address this limitation, recent work has explored rubric evolution~\citep{zhang2026chasing,shen2026rethinking}, drawing on the self-improvement capabilities of LLMs to refine generated rubrics through successive rounds of revision and comparison. 
This approach offers a promising way to obtain high-quality rubrics without requiring human annotation~\citep{mahmoud2026rewardhacking}.

Accurate rubric comparison is central to rubric evolution, since  it determines which revisions are retained and thereby guides the direction of evolution. However, this task is difficult even for human experts, which presents two critical challenges. \textbf{\ding{182} Rubric comparison should reflect downstream task performance.} The practical value of a rubric depends on its ability to improve response quality after RLHF. Yet directly assessing this value by training and evaluating a policy with each candidate rubric is often computationally prohibitive. \textbf{\ding{183} Rubric comparison should discourage unnecessary criteria.} A criterion may appear reasonable but does not improve response quality. Such criteria increase verification costs and may dilute the influence of essential requirements~\citep{mahmoud2026rewardhacking}, which should be identified and discouraged in rubric comparison process.

To address these challenges, we introduce \textbf{Rubrics on Trial}, a multi-agent framework that evolves rubrics through comparisons of synthetic response pairs. To address challenge \ding{182}, it generates and compares responses that satisfy the respective candidate rubrics. This comparison serves as a proxy for assessing each rubric's potential to improve response quality after RLHF, without requiring training a separate policy for every candidate. To address challenge \ding{183}, it generates high-quality alternative responses that violate a candidate criterion, then compares these with edited versions that satisfy the criterion, to assess whether the criterion is necessary for improving response quality. A rubric is favored only when both comparisons favor it, and the comparison result guides rubric evolution.

Our contributions can be summarized as follows. \ding{182} \textbf{We identify two critical challenges in rubric comparison,} which should reflect downstream task performance while discouraging unnecessary criteria. \ding{183} \textbf{We introduce Rubrics on Trial, a multi-agent framework for rubric comparison and evolution.} It addresses the identified challenges through synthetic response pairs and alternative responses, and provides a theoretical guarantee for the surrogate comparison strategy. \ding{184} \textbf{We demonstrate the effectiveness of the proposed method through extensive experiments.} Our method outperforms existing rubric generation approaches, and evolution effectively improves the quality of generated rubrics. These improvements translate into better downstream RLHF performance.

\section{Preliminaries}
\label{sec:preliminaries}

\subsection{Rubric-based Evaluation}
\label{subsec:rubric_rl}

Rubric-based evaluation assesses response quality against explicit criteria~\citep{hashemi2024llmrubric}.
Given a user query $x \in \mathcal{X}$ and an LLM-generated response $y \in \mathcal{Y}$, we define a rubric $r = \{c_1, c_2, \ldots, c_\mathrm{K}\}$ comprising $\mathrm{K}$ readily verifiable criteria. Each criterion is a textual statement characterizing a property of an ideal response.
Suppose $v(\cdot)$ is an LLM-based verifier that assesses whether the response $y$ satisfies the criterion $c_k$ given $x$, the evaluation score is defined as: 
\begin{equation}
    s(x,y)=\frac{1}{\mathrm{K}}\sum_{k=1}^\mathrm{K} v(x,y,c_k),
\end{equation}
which aggregates verification results across all criteria.  $s(x,y)$ provides a comprehensive evaluation of LLM response and supports various downstream applications, such as selecting high-quality responses~\citep{li2026rubrichub}, evaluating LLM agents~\citep{chen2026causalgame}, and providing reward signal in RLHF~\citep{gunjal2026rubrics,pan2026optimal,pan2026uncertainty}.

\subsection{Rubric Generation and Evolution}
\label{subsec:problem_definition}

Rubric generation is a long-standing and  formidable challenge. Some approaches relied on expert annotation~\citep{arora2025healthbench}, which is costly and difficult to scale. To address this limitation, other methods employ LLMs to generate rubrics directly from user queries~\citep{cook2024ticking,wei2025rocketeval}. The premise is that LLMs can induce what an ideal response should encompass to comprehensively address the query. This approach enables efficient rubric construction at scale, with numerous applications in real-world practices~\citep{dhole2026rubricrag}.

The self-improvement capabilities of LLMs render it feasible to evolutionarily optimize the quality of LLM-generated rubrics.
Specifically, let $r_t$ denote the rubric at round $t$, a single evolution step for $r_t$ can be formulated as follows:
\begin{equation}
    \hat{r}_{t+1}=\operatorname{Propose}(x,r_t), \quad \hat{d}_t=\operatorname{Compare}(x,r_t,\hat{r}_{t+1}), \quad r_{t+1}=\operatorname{Update}(x,r_t,\hat{r}_{t+1},\hat{d}_t),
\end{equation}
which starts from generating a candidate rubric $\hat{r}_{t+1}$ as a variant of $r_t$. A binary comparison signal $\hat{d}_t$ is then yielded, where $\hat{d}_t=1$ indicates that $\hat{r}_{t+1}$ is superior to $r_t$, and $\hat{d}_t=0$ otherwise.
The update follows a simple greedy strategy: $r_{t+1}$ retains the original $r_t$ if $\hat{d}_t=0$, and adopts the candidate $\hat{r}_{t+1}$ if $\hat{d}_t=1$. The quality of $r_t$ is expected to improve as $t$ increases. Rubric evolution has recently attracted increasing research attention. \cite{zhang2026chasing} elicits discriminative rubrics by contrasting high-quality, diverse responses in the high-reward region; \cite{shen2026rethinking} recursively decompose coarse criteria into finer-grained ones and filter out redundant criteria; \cite{xu2026sibylsense} alternate rubric-memory refinement with adversarial updates to uncover supplementary criteria.



\section{Methodology}
\label{sec:methodology}

\subsection{Motivation}
\label{sec:motivation}

Accurate rubric comparison is central to evolving LLM-generated rubrics toward higher quality. It determines whether a candidate rubric outperforms the current rubric (i.e., $\hat{r}_{t+1}\succ r_t$), which guides the direction of evolution. Inaccurate rubric comparison can misdirect the evolutionary process and ultimately undermine the quality of LLM-generated rubrics. 
Therefore, accurate comparison is critical to consistently improve rubric quality during evolution.

However, accurate rubric comparison is non-trivial: even human experts may struggle to judge whether one rubric outperforms another. This difficulty stems from two key challenges.
\textbf{\ding{182} Rubric comparison should reflect downstream task performance.} Specifically, rubrics are often used to provide reward signals in RLHF; consequently, a high-quality rubric should lead to improved model responses after RLHF. However, evaluating such downstream performance requires executing the full RLHF pipeline, which is often prohibitively expensive. Therefore, comparing rubrics based on downstream performance is important but difficult.
\textbf{\ding{183} Rubric comparison should discourage unnecessary criteria.} In practice, unnecessary criteria often exist in LLM-generated rubrics~\citep{mahmoud2026rewardhacking}, which refer to criteria not essential for satisfying the user query. They not only increase verification costs, but may also dilute the influence of truly necessary criteria.  Therefore, rubric comparison should favor rubrics that preserve only necessary ones.


\begin{figure}
    \centering
    \captionsetup[subfigure]{font=small,skip=2pt}
    \settoheight{\motivationfigureheight}{\includegraphics[width=0.49\linewidth]{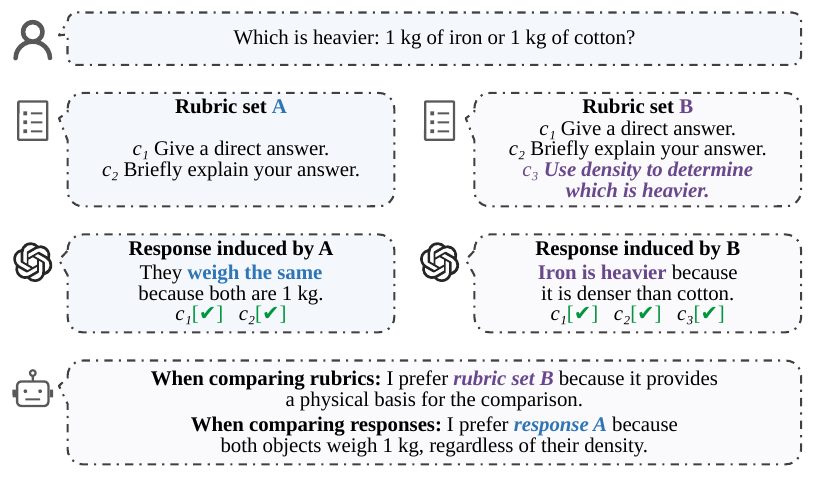}}
    \begin{subfigure}[t]{0.49\linewidth}
        \centering
        \includegraphics[width=\linewidth]{imgs/motivation_challenge1_study_plan_p56.pdf}
        \caption{}
        \label{fig:motivation_study_plan}
    \end{subfigure}%
    \hfill
    \resizebox{!}{\motivationfigureheight}{%
        \begin{tikzpicture}
            \draw[dash pattern=on 14pt off 10.5pt, line width=2pt] (0,0) -- (0,8);
        \end{tikzpicture}%
    }%
    \hfill
    \begin{subfigure}[t]{0.49\linewidth}
        \centering
        \includegraphics[width=\linewidth]{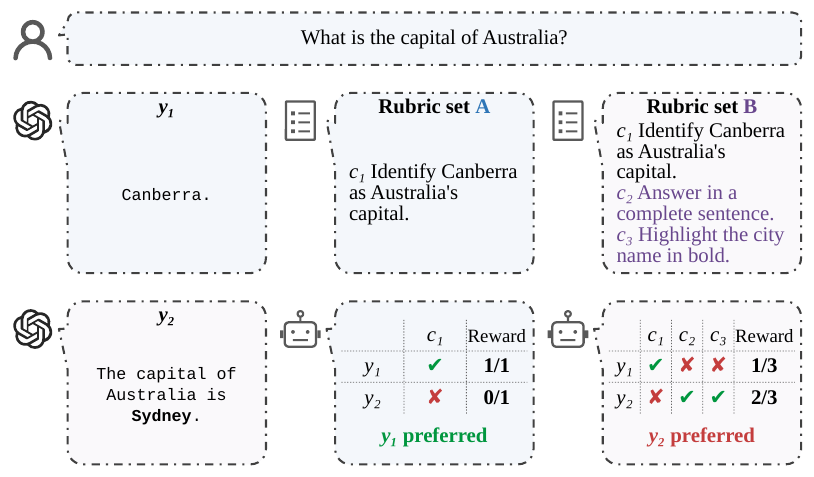}
        \caption{}
        \label{fig:motivation_reward_flip}
    \end{subfigure}
    \caption{Two case studies illustrating challenges in rubric comparison. (a) Direct comparison of rubric texts favors rubric B, although rubric A induces a better response. (b) Unnecessary criteria ($r_2$ and $r_3$) cause the incorrect response ($y_2$) to receive a higher score than the correct response ($y_1$).}
    \label{fig:motivation_case_studies}
    \end{figure}

\textbf{Case studies.} Figure~\ref{fig:motivation_case_studies} illustrates these challenges through two case studies. For challenge \ding{182}, panel (a) presents a scenario in which direct comparison of rubric texts fails to reflect downstream performance. Although text comparison favors rubric B, rubric A induces a better response: it correctly states that both objects weigh 1 kg,
For challenge \ding{183}, panel (b) shows how unnecessary criteria can distort response rankings. Under rubric B, response $y_2$ violates the correctness criterion ($r_1$) but satisfies the style-related criteria ($r_2$ and $r_3$). Consequently, it receives a higher score than $y_1$, which correctly answers ``Canberra'' satisfying the user query but violates the style-related criteria.

Given the importance of accurate rubric comparison for rubric evolution and the challenges outlined above, we seek a rubric comparison strategy that addresses both challenges.
This motivates three research questions: \textit{(1) How can rubric comparison reflect downstream RLHF performance? (2) How can rubric comparison discourage criteria that are unnecessary for satisfying the user query? (3) Does it improve rubric evolution and the effectiveness of the resulting rubrics?}

\subsection{Improving Rubric Comparison via Synthetic Response Pairs}
\label{subsec:synthetic_evidence}

To account for downstream RLHF performance in rubric comparison (Challenge~\ding{182}), a direct approach is to evaluate each rubric through the policy it produces under RLHF.
Given the current rubric $r_t$ and the candidate rubric $\hat{r}_{t+1}$, we independently apply RLHF to the same initial policy $\pi$, using the scores assigned by each rubric as rewards, to obtain policies $\pi_{r_{t}}$ and $\pi_{\hat{r}_{t+1}}$, respectively. For a given query $x$, we evaluate each policy in terms of its expected response quality:
\begin{equation}
\label{eq:post_rl_quality}
    \mu(r_t)=\mathbb E_{y\sim \pi_{{r_t}}(\cdot\mid x)}[g(x,y)], \quad\mu(\hat{r}_{t+1})=\mathbb E_{y\sim \pi_{\hat{r}_{t+1}}(\cdot\mid x)}[g(x,y)],
\end{equation}
where $g(x,y)$ is an independent measure of response quality. In practice, $\mu(r_t)$ can be estimated by sampling responses from $\pi_{r_t}$, scoring them with an LLM judge, and averaging the resulting scores. We prefer $r_t$ to $\hat{r}_{t+1}$, denoted by $r_t \succ \hat{r}_{t+1}$, if $\mu(r_t) > \mu(\hat{r}_{t+1})$, indicating that $r_t$ yields a policy with higher expected response quality after RLHF. However, this approach requires training a separate policy for every candidate rubric, making it prohibitively expensive for rubric comparison.

To reduce computational cost, we propose a surrogate comparison strategy that uses responses sampled from the initial policy $\pi$. Motivated by the expectation that RLHF encourages responses to satisfy the reward rubric, we approximate the post-RLHF response distribution by conditioning $\pi$ on rubric satisfaction. The resulting surrogates for $\mu(r_t)$ and $\mu(\hat{r}_{t+1})$ are:
\begin{equation}
\label{eq:rubric_satisfying_quality}
\hat{\mu}(r_{t})
=
\mathbb{E}_{y\sim\pi(\cdot\mid x)}
\bigl[g(x,y)\mid y\models r_t\bigr],
\qquad \hat{\mu}(\hat{r}_{t+1})
=
\mathbb{E}_{y\sim\pi(\cdot\mid x)}
\bigl[g(x,y)\mid y\models \hat{r}_{t+1}\bigr],
\end{equation}
where \(y\models r\) indicates that \(y\) satisfies all criteria in \(r\). These quantities can be estimated by sampling responses from \(\pi(\cdot\mid x)\) and, for each rubric, averaging the quality scores of the responses that satisfy it. We then prefer $r_t\succ \hat{r}_{t+1}$, if $\hat{\mu}(r_{t})>\hat{\mu}(\hat{r}_{t+1})$. This strategy eliminates the need for RLHF during rubric comparison. Besides, it preserves the comparison preference obtained by evaluating the post-RLHF policies under an idealized training model that fully satisfies each rubric while minimizing KL divergence from the initial policy (see Theorem~\ref{thm:rank_preservation}). Therefore, this strategy provides an efficient approach to rubric comparison that accounts for downstream RLHF performance.

\begin{theorem}
\label{thm:rank_preservation}
Assume that rubric satisfaction has positive probability under the initial policy $\pi$, so that $\hat{\mu}$ is well-defined. For each $r\in\{r_t,\hat r_{t+1}\}$, suppose the post-RLHF policy minimizes KL divergence from $\pi$ among policies that fully satisfy its reward rubric:
\begin{equation}
\label{eq:complete_projection}
\pi_r(\cdot\mid x)
=\operatorname*{arg\,min}_{q:\,\mathbb P_{y\sim q}(y\models r)=1}
\KL\!\left(q\,\Vert\,\pi(\cdot\mid x)\right).
\end{equation}
Then $\pi_r(\cdot\mid x)=\pi(\cdot\mid x,y\models r)$ and $\mu(r)=\hat{\mu}(r)$. Consequently,
\begin{equation}
\label{eq:complete_rank_preservation}
\hat{\mu}(r_t)>\hat{\mu}(\hat r_{t+1})
\quad\Longleftrightarrow\quad
\mu(r_t)>\mu(\hat r_{t+1}).
\end{equation}
\end{theorem}

The implementation of the surrogate comparison strategy is detailed as follows. At round $t$, we construct the candidate by adding a criterion $c$:
\begin{equation}
    \hat r_{t+1}=r_t\cup\{c\}.
\end{equation}
\begin{wraptable}{r}{0.5\linewidth}
\vspace{-8pt}
\centering
\caption{Rubric comparison accuracy.}
\label{tab:motivation_comparison_accuracy}
\setlength{\tabcolsep}{2pt}
\renewcommand{\arraystretch}{1}
\fontsize{8.5}{10.2}\selectfont
\begin{threeparttable}
\def\comparisonrows{
\toprule
\multirow{2.5}{*}{\textbf{Compare}} & \multicolumn{2}{c}{\textbf{RubricBench}} & \multicolumn{2}{c}{\textbf{HealthBench}} & \multicolumn{2}{c}{\textbf{ResearchQA}} \\
\cmidrule(lr){2-3} \cmidrule(lr){4-5} \cmidrule(lr){6-7}
& ACC & $\Delta$ & ACC & $\Delta$ & ACC & $\Delta$ \\
\midrule
\rowcolor[HTML]{f0f0f0}
\multicolumn{7}{l}{\textit{\textbf{Comparator: DeepSeek-V4-Flash}}} \\
\rowcolor{white}
Rubric text & 0.72 & - & 0.40 & - & 0.38 & - \\
\rowcolor[HTML]{ecf0ff}
\textbf{Responses} & \textbf{0.84} & \textbf{16.67\%$\uparrow$} & \textbf{0.91} & \textbf{126.25\%$\uparrow$} & \textbf{0.73} & \textbf{90.79\%$\uparrow$} \\
\midrule
\rowcolor[HTML]{f0f0f0}
\multicolumn{7}{l}{\textit{\textbf{Comparator: Gemini-3.1-Flash-Lite}}} \\
\rowcolor{white}
Rubric text & 0.36 & - & 0.15 & - & 0.34 & - \\
\rowcolor[HTML]{ecf0ff}
\textbf{Responses} & \textbf{0.77} & \textbf{113.89\%$\uparrow$} & \textbf{0.76} & \textbf{420.69\%$\uparrow$} & \textbf{0.65} & \textbf{91.18\%$\uparrow$} \\
\bottomrule
}
\setlength{\tabcolsep}{0pt}
\setbox0=\hbox{\begin{tabular}{lcccccc}\comparisonrows\end{tabular}}
\dimen0=\dimexpr\linewidth-\wd0\relax
\divide\dimen0 by 14
\setlength{\tabcolsep}{\dimen0}
\begin{tabular}{lcccccc}\comparisonrows\end{tabular}
\begin{tablenotes}
\item \textit{Note:} $\Delta$ indicates the relative improvement over rubric-text comparison in percentage. The best results are \textbf{bolded}.
\end{tablenotes}
\end{threeparttable}
\vspace{-6pt}
\end{wraptable}

Under $\pi$, responses satisfying $r_t$ either satisfy or violate $c$. Comparing the expected qualities of these two groups yields the same rubric ranking as the surrogate comparison (Corollary~\ref{cor:add_rank_preservation}). We use a synthetic response pair as a practical proxy for this population comparison.

Given a query $x$, we generate a response $a^+$ that satisfies all criteria in $\hat{r}_{t+1}$. We then minimally edit $a^+$ to obtain $a^-$, which violates $c$ but satisfies all criteria in $r_t$. The minimal editing here aims to isolate the impact of $c$ by removing unrelated changes. To enforce the rubric-satisfaction constraints, we sample multiple responses from $\pi$ and retain those that pass the following checks by a verifier $v$:
\begin{equation}
\label{eq:synthetic_pair_conditions}
\begin{aligned}
v(x,a^+,c)&=1,\qquad v(x,a^-,c)=0,\\
v(x,a^+,c_k)&=1,\qquad v(x,a^-,c_k)=1,
\quad \forall c_k\in r_t.
\end{aligned}
\end{equation}

We then use an LLM comparator to compare $a^+$ and $a^-$ given $x$. If $a^+$ is favored, we derive $\hat{r}_{t+1}\succ r_t$. This strategy provides an efficient approach to rubric comparison while accounting for downstream RLHF performance.

\textbf{Case study.} To investigate whether synthetic responses improve the accuracy of rubric comparison. We use an LLM to add a plausible but harmful criterion to standard rubric in public benchmarks. Two LLMs then identify the better rubric by comparing either the rubric texts or their corresponding synthetic responses (see Appendix~\ref{app:implementation} for details).
As shown in Table~\ref{tab:motivation_comparison_accuracy}, response comparison identifies the better rubric more accurately across both LLMs and all three benchmarks, supporting the use of synthetic responses for rubric comparison.

\subsection{Discouraging Unnecessary Criteria with Alternative Responses}
\label{subsec:alternative_comparison}
\begin{figure}[t]
\centering
\includegraphics[width=\linewidth]{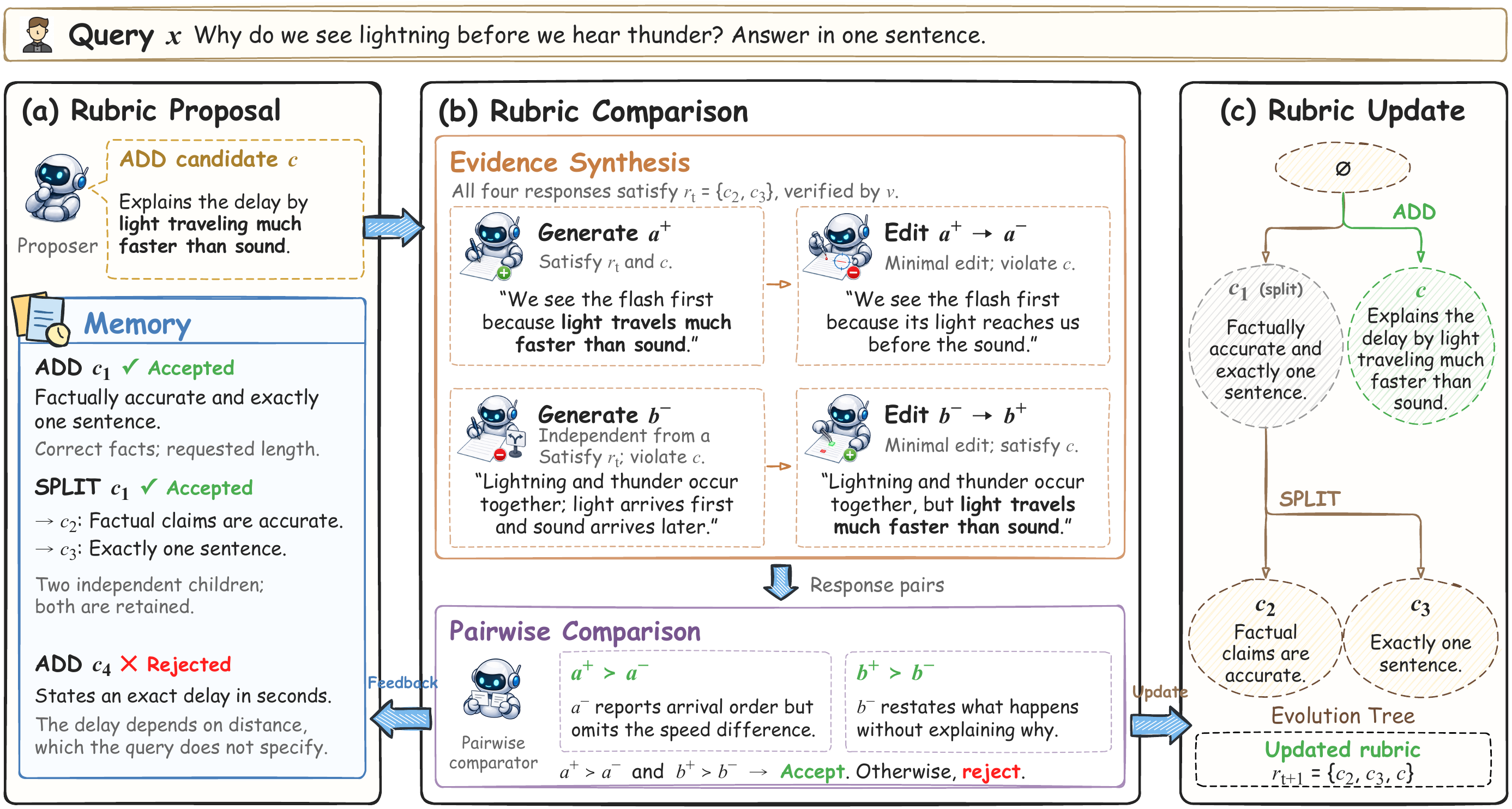}
\caption{The overview of Rubrics on Trial for one round of evolution.}
\label{fig:method_overview}
\end{figure}

Although comparing $a^+$ and $a^-$ accounts for downstream RLHF performance, it cannot reliably discourage unnecessary criteria (Challenge~\ding{183}).
The reason is that $a^-$ is constructed by editing $a^+$ specifically to violate $c$. In practice, such edits often remove  the content associated with $c$ entirely, producing a degraded response rather than a response that satisfies the user query in another way.
Consequently, the comparator may favor $a^+$ even when $c$ is unnecessary to satisfy the user query.

More specifically, consider the query ``Which is a prime number? (A) 21 (B) 27 (C) 29'' with the current rubric $r_t=\{c_1\}$, where $c_1$ is ``Select (C) 29.''
Suppose the additional criterion $c$ is ``Check the options in at least two separate reasoning steps.''
The response $a^+$ satisfies $c$ by first showing that 21 and 27 are composite through factorization and then establishing that 29 is prime. Editing $a^+$ to violate $c$ removes both reasoning steps, leaving only ``Answer: (C)'' in $a^-$.
The comparator favors $a^+$ because $a^-$ gives no justification.
However, $c$ is unnecessary: a single-step response such as ``29 is the only option not divisible by 3, so the answer is (C)'' also provides a valid justification for selecting the correct option. 
Thus, the comparator may favor $a^+$ when $c$ is an unnecessary criterion.

To address this limitation, we introduce comparison with alternative responses.
The key idea is that an unnecessary criterion can be violated with little or no loss in response quality, since the user query can be satisfied through an alternative approach. We therefore test whether satisfying $c$ improves the quality of a strong response independently generated to violate $c$.
Specifically, given a query $x$, we generate an alternative response $b^-$ of the highest possible quality, subject to violating $c$ while satisfying all criteria in $r_t$. Unlike $a^-$, $b^-$ is generated from scratch rather than edited from a response constructed to satisfy $c$. This allows $b^-$ to address the user query bypassing $c$. We then minimally edit $b^-$ to obtain $b^+$, which satisfies both $c$ and $r_t$. To enforce rubric-satisfaction constraints, we retain responses from $\pi$ that pass the following checks:
\begin{equation}
\label{eq:alternative_pair_conditions}
\begin{aligned}
v(x,b^+,c)&=1,\qquad v(x,b^-,c)=0,\\
v(x,b^+,c_k)&=1,\qquad v(x,b^-,c_k)=1,
\quad \forall c_k\in r_t.
\end{aligned}
\end{equation}

We then use the LLM comparator to assess the relative quality of $b^+$ and $b^-$ given $x$. A preference for $b^+$ suggests that satisfying $c$ improves even a strong alternative response, providing evidence for the necessity of $c$. Conversely, a tie or a preference for $b^-$ suggests that the user query can be satisfied at least as well without $c$, providing evidence that $c$ is unnecessary.
In the above example, the single-step response serves as $b^-$, and editing $b^-$ to check the options in two separate steps yields $b^+$.
Since both responses select (C) with a correct justification, the comparator finds $b^+$ and $b^-$ equally good, so $c$ is identified as unnecessary.
The complete case is presented in Appendix~\ref{app:alternative_case}.

Combining the two comparison strategies, we define the rubric comparison signal as
\begin{equation}
\label{eq:rubric_comparison_signal}
\hat d_t
=\operatorname{Compare}(x,r_t,\hat r_{t+1})
=\begin{cases}
1, & a^+\succ a^- \ \text{and}\ b^+\succ b^-,\\
0, & \text{otherwise},
\end{cases}
\end{equation}
where $\hat d_t=1$ only when $\hat r_{t+1}$ improves response quality in both comparisons. This comparison signal reflects downstream RLHF performance while discouraging unnecessary criteria.

\subsection{The Workflow of Rubrics on Trial}
\label{subsec:workflow}

In this section, we introduce Rubrics on Trial, a multi-agent framework that uses the comparison signal in (\ref{eq:rubric_comparison_signal}) to guide rubric evolution. As illustrated in Figure~\ref{fig:method_overview}, the framework comprises a rubric proposer, response generators, a verifier, and a pairwise comparator. These agents iteratively refine rubrics through the proposal, comparison, and update procedure introduced in Section~\ref{subsec:problem_definition}. Algorithm~\ref{alg:rubric_evolution} summarizes the workflow, with the core steps described below.

\Needspace{20\baselineskip}
\begingroup
\setlength{\intextsep}{0pt}
\begin{wrapfloat}{algorithm}{r}{0.5\linewidth}
\captionsetup{font=footnotesize}
\caption{The workflow of Rubrics on Trial.}
\label{alg:rubric_evolution}
\footnotesize
{\raggedright
\textbf{Input}: $x$: user query; $T$: maximum evolution rounds; $P$: patience.\par
\textbf{Output}: $r^*$: evolved rubric.\par}
\setboolean{ALC@noend}{true}
\renewcommand{\algorithmiccomment}[1]{#1}
\begin{algorithmic}[1]
\STATE $r_0\leftarrow\varnothing$; $\mathcal H\leftarrow\varnothing$; $u\leftarrow0$
\FOR{$t=0,\ldots,T-1$}
    \STATE $\hat r_{t+1}\leftarrow\operatorname{Propose}(x,r_t,\mathcal{H})$
    \IF{ADD}
        \STATE $\hat d_t\leftarrow\operatorname{Compare}(x,r_t,\hat r_{t+1})$
        \STATE $r_{t+1}\leftarrow\operatorname{Update}(x,r_t,\hat r_{t+1},\hat d_t)$
    \ELSIF[$r_{t+1}\leftarrow\varnothing$]{SPLIT}
        \FORALL{children $c_i$ of $c_p$}
            \STATE $\hat d_t^{(i)}\leftarrow\operatorname{Compare}\!\bigl(x,r_t\setminus\{c_p\},\hat r_{t+1}^{(i)}\bigr)$
            \STATE $\begin{aligned}[t]
                &r_{t+1}\leftarrow r_{t+1}\cup{}\\[-3pt]
                &\quad\operatorname{Update}\!\bigl(x,r_t\setminus\{c_p\},\hat r_{t+1}^{(i)},\hat d_t^{(i)}\bigr)
            \end{aligned}$
        \ENDFOR
    \ENDIF
    \STATE $\mathcal H\leftarrow\mathcal H\cup\{(\hat r_{t+1},r_{t+1},\text{feedback})\}$
    \STATE $u\leftarrow u+1$ \textbf{if} $r_{t+1}=r_t\ne\varnothing$, \textbf{else} $0$
    \IF{$u\ge P$}
        \STATE \textbf{break}
    \ENDIF
\ENDFOR
\STATE  $r^*\leftarrow r_{t+1}$
\end{algorithmic}
\end{wrapfloat}

\textbf{Rubric proposal.}
Starting from an empty rubric $r_0=\varnothing$, the proposer uses the query, the current rubric, and memory $\mathcal H$ to propose an ADD or SPLIT operation.
ADD introduces an additional criterion $c$, while SPLIT decomposes a criterion $c_p$ into finer criteria $c_i$.

\textbf{Rubric comparison and update.}
For ADD, we compare $r_t$ and $\hat{r}_{t+1}$ using the comparison signal $\hat d_t$ defined in (\ref{eq:rubric_comparison_signal}). If $\hat d_t=1$, we set $r_{t+1}=\hat r_{t+1}$; otherwise, we retain $r_t$.
For SPLIT, we first remove the parent criterion $c_p$ to obtain a shared baseline $r_t\setminus{c_p}$. Each child criterion $c_i$ defines a candidate rubric
\[
\hat r_{t+1}^{(i)}
=(r_t\setminus\{c_p\})\cup\{c_i\}.
\]
We then compare each candidate independently against the shared baseline to obtain the corresponding signal $\hat d_t^{(i)}$. A child criterion $c_i$ is accepted if $\hat d_t^{(i)}=1$. The updated rubric $r_{t+1}$ consists of the baseline criteria and all accepted child criteria.

The memory $\mathcal H$ stores historical information to guide subsequent proposals.
These steps are iterated for $T$ rounds, stopping early when a nonempty rubric remains unchanged for $P$ consecutive rounds.

\endgroup

\Needspace{17\baselineskip}
\section{Experiments}
\label{sec:experiments}
\begingroup
\raggedbottom
\renewcommand{\topfraction}{0.9}
\renewcommand{\textfraction}{0.07}
\renewcommand{\floatpagefraction}{0.85}
\setlength{\textfloatsep}{11pt plus 2pt minus 2pt}
\setlength{\floatsep}{10pt plus 2pt minus 2pt}
\setlength{\intextsep}{11pt plus 2pt minus 2pt}


\subsection{Experimental Setup}
\label{subsec:experimental_setup}

\textbf{Datasets and Evaluation.}
We conduct empirical evaluations on JudgeBench~\citep{tan2025judgebench}, RM-Bench~\citep{liu2024rmbench}, and RewardBench Chat-Hard~\citep{lambert2024rewardbench}, covering nine evaluation sets.
We use each generated rubric to rank benchmark responses by their evaluation scores $s(x,y)$.
We report preference accuracy, with half credit for ties, and its unweighted mean across the nine sets (Macro$_{9}$).
Higher accuracy indicates better discrimination of response quality.

\textbf{Baselines.}
We compare against two groups of methods:
(1) \textbf{LLM-based rubric generation methods}, which prompt fixed LLMs to generate rubrics without training, including the single-pass methods direct generation (Direct), TICK~\citep{cook2024ticking}, and RocketEval~\citep{wei2025rocketeval}, and the rubric evolution methods RRD-Uniform~\citep{shen2026rethinking} and ChaseTail~\citep{zhang2026chasing}; and
(2) \textbf{Trained rubric generators}, which are dedicated models trained to generate rubrics, including RubricRM~\citep{liu2025openrubrics}, RubricARM~\citep{xu2026rubricarm}, and RubricARROW~\citep{jiang2026rubricarrow}.
All methods generate rubrics from the query without access to the benchmark responses.

\textbf{Implementation Details.}
Unless otherwise specified, we use DeepSeek-V4-Flash for rubric generation, response generation and editing, and pairwise comparison.
All methods use GPT-4.1-mini as the final verifier and uniform criterion weights to isolate rubric quality.
Rubrics on Trial runs for at most $T=10$ rounds with early stopping patience $P=5$.
Further details appear in Appendix~\ref{app:experimental_details}.

\begin{table}
\caption{Pairwise preference accuracy (\%) across nine evaluation sets from three benchmark suites.}
\label{tab:main_results}
\centering
\renewcommand{\arraystretch}{1}
\setlength{\tabcolsep}{2.8pt}
\fontsize{8.4}{10.2}\selectfont
\setlength{\dashlinedash}{2.2pt}\setlength{\dashlinegap}{2.6pt}
\def\experimentrows{
\toprule
 & \multicolumn{4}{c}{\textbf{JudgeBench}} & \multicolumn{4}{c}{\textbf{RM-Bench}} & \textbf{RewardBench} & \textbf{Macro} \\
\cmidrule(lr){2-5}\cmidrule(lr){6-9}\cmidrule(lr){10-10}\cmidrule(lr){11-11}
\textbf{Method} & Knowledge & Reasoning & Math & Coding & Chat & Math & Code & Safety & Chat-Hard & Avg. \\
\midrule
\rowcolor{experimentgray}\multicolumn{11}{l}{\textbf{\textit{LLM-based rubric generation methods}}} \\
\rowcolor{white}
Direct & 67.53 & 62.08 & 75.56 & 67.12 & 63.01 & 79.63 & 65.40 & 86.09 & 73.25 & 71.07 \\
\rowcolor{white}
TICK & 67.86 & 62.75 & 72.78 & 68.49 & 62.96 & 82.52 & 66.93 & 87.69 & 72.81 & 71.64 \\
\rowcolor{white}
RocketEval & 69.97 & 70.47 & 77.22 & 68.49 & 63.48 & 83.29 & 66.98 & 87.55 & 71.93 & 73.26 \\
\rowcolor{white}
RRD-Uniform & 72.56 & 72.82 & 82.22 & 66.44 & 58.83 & 81.06 & 66.86 & 76.37 & 66.78 & 71.55 \\
\rowcolor{white}
ChaseTail & 71.59 & 84.23 & 82.22 & \textbf{69.86} & 63.05 & 85.55 & 68.57 & 87.02 & 72.04 & 76.01 \\
\hdashline
\rowcolor{experimentgray}\multicolumn{11}{l}{\textbf{\textit{Trained rubric generators}}} \\
\rowcolor{white}
RubricRM & 58.60 & 58.05 & 70.00 & 66.44 & 65.37 & 67.07 & 69.20 & 76.73 & 71.60 & 67.01 \\
\rowcolor{white}
RubricARM & 57.95 & 58.72 & 67.78 & 68.49 & 66.32 & 67.31 & 69.13 & 75.90 & \textbf{77.95} & 67.73 \\
\rowcolor{white}
RubricARROW & 53.90 & 57.38 & 66.11 & 67.81 & 66.75 & 63.97 & 65.91 & 67.90 & 75.99 & 65.08 \\
\hdashline
\rowcolor{experimentours}
\textbf{Rubrics on Trial} & \textbf{74.03} & \textbf{89.60} & \textbf{85.56} & 68.49 & \textbf{67.96} & \textbf{91.27} & \textbf{69.66} & \textbf{89.64} & 76.97 & \textbf{79.24} \\
\bottomrule
}
\setlength{\tabcolsep}{0pt}
\setbox0=\hbox{\begin{tabular}{l*{10}{c}}\experimentrows\end{tabular}}
\dimen0=\dimexpr\linewidth-\wd0\relax
\divide\dimen0 by 22
\setlength{\tabcolsep}{\dimen0}
\begin{tabular}{l*{10}{c}}\experimentrows\end{tabular}
\par\vspace{3pt}\raggedright\scriptsize
\textit{Note:} Macro Avg. denotes the mean across all evaluation sets. The best metrics in each column are \textbf{bolded}.
\end{table}

\subsection{Overall Performance}
\label{subsec:overall_performance}

In this section, we compare Rubrics on Trial with competitive LLM-based rubric generation methods and trained rubric generators.
The results are presented in Table~\ref{tab:main_results} with key observations as follows.
\textbf{\ding{182} Single-pass rubric generation offers limited gains.}
TICK and RocketEval improve only slightly over Direct.
This indicates that generating rubrics in a single pass, even with carefully designed prompts, brings limited improvement in rubric quality.
\textbf{\ding{183} Rubric evolution is effective to improve rubric quality.}
For example, ChaseTail achieves the strongest baseline performance. However, RRD-Uniform barely improves over Direct.
This suggests that rubric evolution is promising, but its gains remain unstable without accurate rubric comparison.
\textbf{\ding{184} Trained rubric generators transfer unevenly across evaluation sets.}
They perform well on chat-oriented evaluation sets such as RM-Bench Chat, but underperform Direct on most other sets, leading to a lower Macro$_{9}$ than Direct.
This suggests that, despite the training cost, a dedicated rubric generator does not ensure rubric quality across evaluation sets.
\textbf{\ding{185} Rubrics on Trial achieves the best rubric quality.}
It attains the highest Macro$_{9}$ and ranks first on seven of nine evaluation sets, outperforming Direct and the strongest baseline, ChaseTail, by $8.17$ and $3.23$ percentage points, respectively.
This success is attributed to its comparison strategy, which facilitates evolution and improves the quality of generated rubrics.

\subsection{Ablation Study}
\label{subsec:ablation_study}

In this section, we dissect the individual contributions of the components within Rubrics on Trial.
In Table~\ref{tab:ablation_evolution}, Rubrics on Trial$^{\dagger}$ replaces iterative rubric evolution with a single revision round, in which the proposer may modify any criteria in the rubric generated by Direct. Rubrics on Trial$^{\ddagger}$ retains rubric evolution but compares rubric texts rather than synthetic response pairs.
The key observations are as follows.
\textbf{\ding{182} Rubric evolution benefits the quality of rubrics.}
Rubrics on Trial$^{\dagger}$ underperforms the full method on all nine evaluation sets (e.g., Macro$_{9}$ drops from $79.24\%$ to $73.41\%$), validating the importance of multiple rounds of proposal and comparison.
\textbf{\ding{183} Synthetic response pairs improve the performance of rubric comparison.}
Rubrics on Trial$^{\ddagger}$ underperforms the full method on eight of nine evaluation sets (e.g., Macro$_{9}$ drops to $74.28\%$), validating the importance of comparing rubrics through the responses that satisfy them.

In Table~\ref{tab:ablation_pairs}, Rubrics on Trial$^{\dagger}$ and Rubrics on Trial$^{\ddagger}$ remove $(a^+,a^-)$ and $(b^+,b^-)$, respectively, and compare rubrics with the remaining pair alone.
The key observations are as follows.
\textbf{\ding{182} Alternative responses help discourage unnecessary criteria.}
Rubrics on Trial$^{\ddagger}$ underperforms the full method on eight of nine evaluation sets (e.g., Macro$_{9}$ drops from $79.24\%$ to $76.11\%$), validating the importance of testing whether an additional criterion improves a strong response that violates it.
\textbf{\ding{183} The two response pairs provide complementary benefits.}
The full method outperforms both variants on eight of nine evaluation sets, showing that neither pair alone suffices.

\begin{table}[!t]
\caption{Ablation study on rubric evolution and synthetic response pairs.}
\label{tab:ablation_evolution}
\centering
\renewcommand{\arraystretch}{0.9}
\fontsize{8.1}{9.8}\selectfont
\def\experimentrows{
\toprule
\multirow{2}{*}{\textbf{Method}} & \multirow{2}{*}{\textbf{Evol.}} & \multirow{2}{*}{\textbf{Resp.}} & \multicolumn{4}{c}{\textbf{JudgeBench}} & \multicolumn{4}{c}{\textbf{RM-Bench}} & \textbf{Reward} & \textbf{Macro} \\
\cmidrule(lr){4-7}\cmidrule(lr){8-11}\cmidrule(lr){12-12}\cmidrule(lr){13-13}
 & & & Know. & Reason. & Math & Code & Chat & Math & Code & Safety & Chat-Hard & Avg. \\
\midrule
\rowcolor{white}
Direct & \XSolidBrush & \XSolidBrush & 67.53 & 62.08 & 75.56 & 67.12 & 63.01 & 79.63 & 65.40 & 86.09 & 73.25 & 71.07 \\
\rowcolor{white}
Rubrics on Trial$^{\dagger}$ & \XSolidBrush & \Checkmark & 69.32 & 70.47 & 78.89 & 67.81 & 62.62 & \underline{84.32} & 66.76 & \underline{86.53} & \underline{74.01} & 73.41 \\
\rowcolor{white}
Rubrics on Trial$^{\ddagger}$ & \Checkmark & \XSolidBrush & \underline{70.78} & \underline{76.17} & \underline{80.56} & \textbf{69.18} & \underline{63.09} & 82.12 & \underline{67.45} & 86.17 & 73.03 & \underline{74.28} \\
\rowcolor{experimentours}
\textbf{Rubrics on Trial} & \Checkmark & \Checkmark & \textbf{74.03} & \textbf{89.60} & \textbf{85.56} & \underline{68.49} & \textbf{67.96} & \textbf{91.27} & \textbf{69.66} & \textbf{89.64} & \textbf{76.97} & \textbf{79.24} \\
\bottomrule
}
\setlength{\tabcolsep}{0pt}
\setbox0=\hbox{\begin{tabular}{l*{12}{c}}\experimentrows\end{tabular}}
\dimen0=\dimexpr\linewidth-\wd0\relax
\divide\dimen0 by 26
\setlength{\tabcolsep}{\dimen0}
\begin{tabular}{l*{12}{c}}\experimentrows\end{tabular}
\par\vspace{3pt}\raggedright\scriptsize
\textit{Note:} Macro Avg. denotes the mean across all evaluation sets. The best and second-best metrics  are \textbf{bolded} and \underline{underlined}, respectively. 
\end{table}

\begin{table}[!t]
\caption{Ablation study on the two response pairs $(a^+,a^-)$ and $(b^+,b^-)$.}
\label{tab:ablation_pairs}
\centering
\renewcommand{\arraystretch}{1.0}
\fontsize{8.1}{9.8}\selectfont
\def\experimentrows{
\toprule
\multirow{2}{*}{\textbf{Method}} & \multirow{2}{*}{$(a^+,a^-)$} & \multirow{2}{*}{$(b^+,b^-)$} & \multicolumn{4}{c}{\textbf{JudgeBench}} & \multicolumn{4}{c}{\textbf{RM-Bench}} & \textbf{Reward} & \textbf{Macro} \\
\cmidrule(lr){4-7}\cmidrule(lr){8-11}\cmidrule(lr){12-12}\cmidrule(lr){13-13}
 & & & Know. & Reason. & Math & Code & Chat & Math & Code & Safety & Chat-Hard & Avg. \\
\midrule
\rowcolor{white}
Direct & \XSolidBrush & \XSolidBrush & 67.53 & 62.08 & 75.56 & 67.12 & 63.01 & 79.63 & 65.40 & 86.09 & 73.25 & 71.07 \\
\rowcolor{white}
Rubrics on Trial$^{\dagger}$ & \XSolidBrush & \Checkmark & \underline{72.08} & \underline{87.58} & \underline{81.67} & \textbf{70.55} & \underline{65.03} & \underline{87.62} & \underline{69.03} & 87.20 & 74.45 & \underline{77.25} \\
\rowcolor{white}
Rubrics on Trial$^{\ddagger}$ & \Checkmark & \XSolidBrush & 69.16 & 85.91 & 79.44 & \underline{69.86} & 63.91 & 85.34 & 68.47 & \underline{87.99} & \underline{74.89} & 76.11 \\
\rowcolor{experimentours}
\textbf{Rubrics on Trial} & \Checkmark & \Checkmark & \textbf{74.03} & \textbf{89.60} & \textbf{85.56} & 68.49 & \textbf{67.96} & \textbf{91.27} & \textbf{69.66} & \textbf{89.64} & \textbf{76.97} & \textbf{79.24} \\
\bottomrule
}
\setlength{\tabcolsep}{0pt}
\setbox0=\hbox{\begin{tabular}{l*{12}{c}}\experimentrows\end{tabular}}
\dimen0=\dimexpr\linewidth-\wd0\relax
\divide\dimen0 by 26
\setlength{\tabcolsep}{\dimen0}
\begin{tabular}{l*{12}{c}}\experimentrows\end{tabular}
\par\vspace{3pt}\raggedright\scriptsize
\textit{Note:} Macro Avg. denotes the mean across all evaluation sets. The best and second-best metrics  are \textbf{bolded} and \underline{underlined}, respectively. 
\end{table}

\Needspace{18\baselineskip}
\subsection{Generalization across LLMs}
\label{subsec:llm_generalization}

\begin{wraptable}{r}{0.5\textwidth}
\vspace{-15pt}
\centering
\captionsetup{font=small,aboveskip=0pt,belowskip=4pt}
\caption{Performance across generation LLMs.}
\label{tab:generalization}
\renewcommand{\arraystretch}{1.03}
\fontsize{7.5}{8.5}\selectfont
\def\experimentrows{
\toprule
\multirow{2}{*}{\textbf{Method}} & \multicolumn{2}{c}{\textbf{JudgeBench}} & \multicolumn{2}{c}{\textbf{RM-Bench}} & \multicolumn{2}{c}{\textbf{Chat-Hard}} \\
\cmidrule(lr){2-3}\cmidrule(lr){4-5}\cmidrule(lr){6-7}
 & ACC & $\Delta$ & ACC & $\Delta$ & ACC & $\Delta$ \\
\midrule
\rowcolor{experimentgray}\multicolumn{7}{l}{\textbf{\textit{Generation LLM: Gemini-3.5-Flash-Lite}}} \\
\rowcolor{white}
Direct & 65.08 & - & 74.89 & - & 69.08 & - \\
\rowcolor{experimentours}
\textbf{Rubrics on Trial} & \textbf{75.97} & 16.73\%$\uparrow$ & \textbf{81.99} & 9.48\%$\uparrow$ & \textbf{73.36} & 6.20\%$\uparrow$ \\
\midrule
\rowcolor{experimentgray}\multicolumn{7}{l}{\textbf{\textit{Generation LLM: GLM-5.1}}} \\
\rowcolor{white}
Direct & 78.71 & - & 79.16 & - & 72.15 & - \\
\rowcolor{experimentours}
\textbf{Rubrics on Trial} & \textbf{86.21} & 9.53\%$\uparrow$ & \textbf{83.37} & 5.32\%$\uparrow$ & \textbf{74.78} & 3.65\%$\uparrow$ \\
\midrule
\rowcolor{experimentgray}\multicolumn{7}{l}{\textbf{\textit{Generation LLM: Qwen3.5-397B-A17B}}} \\
\rowcolor{white}
Direct & 71.21 & - & 73.08 & - & 74.01 & - \\
\rowcolor{experimentours}
\textbf{Rubrics on Trial} & \textbf{81.45} & 14.38\%$\uparrow$ & \textbf{78.79} & 7.81\%$\uparrow$ & \textbf{75.99} & 2.68\%$\uparrow$ \\
\midrule
\rowcolor{experimentgray}\multicolumn{7}{l}{\textbf{\textit{Generation LLM: GPT-4.1}}} \\
\rowcolor{white}
Direct & 60.89 & - & 72.13 & - & 70.72 & - \\
\rowcolor{experimentours}
\textbf{Rubrics on Trial} & \textbf{63.95} & 5.03\%$\uparrow$ & \textbf{77.61} & 7.60\%$\uparrow$ & \textbf{73.57} & 4.03\%$\uparrow$ \\
\bottomrule
}
\setlength{\tabcolsep}{0pt}
\setbox0=\hbox{\begin{tabular}{l*{6}{c}}\experimentrows\end{tabular}}
\dimen0=\dimexpr\linewidth-\wd0\relax
\divide\dimen0 by 14
\setlength{\tabcolsep}{\dimen0}
\begin{tabular}{l*{6}{c}}\experimentrows\end{tabular}
\par\vspace{3pt}\raggedright\scriptsize
\textit{Note:} $\Delta$ indicates the relative improvement over the Direct method in percentage. The best results in each column are \textbf{bolded}.
\vspace{-6pt}
\end{wraptable}

In this section, we evaluate the generalizability of Rubrics on Trial across four generation LLMs from different model families, namely Gemini-3.5-Flash-Lite, GLM-5.1, Qwen3.5-397B-A17B, and GPT-4.1.
For each generation LLM, all agents in the multi-agent system and Direct use that LLM, while GPT-4.1-mini remains the final verifier.
The results are presented in Table~\ref{tab:generalization}, where JudgeBench and RM-Bench report micro-averaged accuracy over their four evaluation sets.
Rubrics on Trial improves over Direct on JudgeBench, RM-Bench, and Chat-Hard for every generation LLM, with relative gains ranging from $2.68\%$ to $16.73\%$.
These consistent gains across model families indicate that the effectiveness of Rubrics on Trial is not tied to a particular generation LLM.

\subsection{Downstream RLHF Performance}
\label{subsec:downstream_rl}

In this section, we investigate whether improved rubrics benefit downstream RLHF performance.
We select strong rubric generation methods from Table~\ref{tab:main_results} as baselines, fine-tune three policies with GRPO~\citep{shao2024deepseekmath} on WildChat~\citep{zhao2024wildchat}, using GPT-4.1-mini as verifier.
GPT-4.1 evaluates the resulting policies on BiGGen-Bench~\citep{kim2025biggen}, AlpacaEval~2.0~\citep{dubois2024length}, and Arena-Hard~\citep{li2024arenahard}.
The results are presented in Table~\ref{tab:downstream_rl}. Overall, rubric-based rewards consistently improve the original policies, which demonstrates the utility of rubric-based rewards for policy optimization. Moreover, Rubrics on Trial achieves the best RLHF performance, indicating that its improved rubric quality translates into better downstream performance.


\begin{table}[!ht]
\caption{Downstream RLHF results.}
\label{tab:downstream_rl}
\centering
\renewcommand{\arraystretch}{0.95}
\fontsize{8.3}{10}\selectfont
\def\experimentrows{
\toprule
\multirow{3}{*}{\textbf{Method}} & \multicolumn{2}{c}{\textbf{BiGGen-Bench}} & \multicolumn{4}{c}{\textbf{AlpacaEval 2.0}} & \multicolumn{4}{c}{\textbf{Arena-Hard}} \\
\cmidrule(lr){2-3}\cmidrule(lr){4-7}\cmidrule(lr){8-11}
 & \multicolumn{2}{c}{Score} & \multicolumn{2}{c}{Vanilla} & \multicolumn{2}{c}{LC} & \multicolumn{2}{c}{Vanilla} & \multicolumn{2}{c}{SC} \\
\cmidrule(lr){2-3}\cmidrule(lr){4-5}\cmidrule(lr){6-7}\cmidrule(lr){8-9}\cmidrule(lr){10-11}
 & Value & $\Delta$ & Value & $\Delta$ & Value & $\Delta$ & Value & $\Delta$ & Value & $\Delta$ \\
\midrule
\rowcolor{experimentgray}\multicolumn{11}{l}{\textbf{\textit{Policy Model: Qwen2.5-3B-Instruct}}} \\
\rowcolor{white}
Original & 64.22 & - & 17.25 & - & 16.71 & - & 25.93 & - & 27.63 & - \\
\rowcolor{white}
Direct & 64.43 & 0.33\%$\uparrow$ & 23.95 & 38.84\%$\uparrow$ & 18.77 & 12.33\%$\uparrow$ & 28.88 & 11.38\%$\uparrow$ & 29.50 & 6.77\%$\uparrow$ \\
\rowcolor{white}
TICK & 65.33 & 1.73\%$\uparrow$ & 20.19 & 17.04\%$\uparrow$ & 18.12 & 8.44\%$\uparrow$ & 27.83 & 7.33\%$\uparrow$ & 29.57 & 7.02\%$\uparrow$ \\
\rowcolor{white}
RocketEval & 64.86 & 1.00\%$\uparrow$ & 24.01 & 39.19\%$\uparrow$ & 18.41 & 10.17\%$\uparrow$ & 31.67 & 22.14\%$\uparrow$ & 30.91 & 11.87\%$\uparrow$ \\
\rowcolor{white}
ChaseTail & 65.90 & 2.62\%$\uparrow$ & 24.13 & 39.88\%$\uparrow$ & 19.43 & 16.28\%$\uparrow$ & 31.73 & 22.37\%$\uparrow$ & 31.94 & 15.60\%$\uparrow$ \\
\rowcolor{experimentours}
\textbf{Rubrics on Trial} & \textbf{66.58} & \textbf{3.67\%$\uparrow$} & \textbf{25.77} & \textbf{49.39\%$\uparrow$} & \textbf{20.56} & \textbf{23.04\%$\uparrow$} & \textbf{32.07} & \textbf{23.68\%$\uparrow$} & \textbf{33.12} & \textbf{19.87\%$\uparrow$} \\
\midrule
\rowcolor{experimentgray}\multicolumn{11}{l}{\textbf{\textit{Policy Model: Qwen2.5-7B-Instruct}}} \\
\rowcolor{white}
Original & 68.98 & - & 27.79 & - & 29.08 & - & 51.90 & - & 53.41 & - \\
\rowcolor{white}
Direct & 70.83 & 2.68\%$\uparrow$ & 35.53 & 27.85\%$\uparrow$ & 31.86 & 9.56\%$\uparrow$ & 54.05 & 4.14\%$\uparrow$ & 55.22 & 3.39\%$\uparrow$ \\
\rowcolor{white}
TICK & 70.50 & 2.20\%$\uparrow$ & 33.84 & 21.77\%$\uparrow$ & 29.63 & 1.89\%$\uparrow$ & 53.90 & 3.85\%$\uparrow$ & 55.49 & 3.89\%$\uparrow$ \\
\rowcolor{white}
RocketEval & 71.58 & 3.77\%$\uparrow$ & 42.14 & 51.64\%$\uparrow$ & 31.89 & 9.66\%$\uparrow$ & 55.97 & 7.84\%$\uparrow$ & 56.09 & 5.02\%$\uparrow$ \\
\rowcolor{white}
ChaseTail & \textbf{71.70} & \textbf{3.94\%$\uparrow$} & 46.39 & 66.93\%$\uparrow$ & 35.73 & 22.87\%$\uparrow$ & 57.40 & 10.60\%$\uparrow$ & 55.83 & 4.53\%$\uparrow$ \\
\rowcolor{experimentours}
\textbf{Rubrics on Trial} & 71.61 & 3.81\%$\uparrow$ & \textbf{47.07} & \textbf{69.38\%$\uparrow$} & \textbf{38.54} & \textbf{32.53\%$\uparrow$} & \textbf{58.56} & \textbf{12.83\%$\uparrow$} & \textbf{57.89} & \textbf{8.39\%$\uparrow$} \\
\midrule
\rowcolor{experimentgray}\multicolumn{11}{l}{\textbf{\textit{Policy Model: Qwen3-4B}}} \\
\rowcolor{white}
Original & 77.05 & - & 50.58 & - & 40.18 & - & 73.04 & - & 72.16 & - \\
\rowcolor{white}
Direct & 77.44 & 0.51\%$\uparrow$ & 52.75 & 4.29\%$\uparrow$ & 42.16 & 4.93\%$\uparrow$ & 74.64 & 2.19\%$\uparrow$ & 73.56 & 1.94\%$\uparrow$ \\
\rowcolor{white}
TICK & 77.92 & 1.13\%$\uparrow$ & 51.85 & 2.51\%$\uparrow$ & 42.84 & 6.62\%$\uparrow$ & 73.49 & 0.62\%$\uparrow$ & 72.43 & 0.37\%$\uparrow$ \\
\rowcolor{white}
RocketEval & 78.18 & 1.47\%$\uparrow$ & \textbf{54.61} & \textbf{7.97\%$\uparrow$} & 43.86 & 9.16\%$\uparrow$ & 74.30 & 1.73\%$\uparrow$ & 72.82 & 0.91\%$\uparrow$ \\
\rowcolor{white}
ChaseTail & 78.88 & 2.38\%$\uparrow$ & 54.29 & 7.33\%$\uparrow$ & 43.71 & 8.79\%$\uparrow$ & 75.18 & 2.93\%$\uparrow$ & 73.67 & 2.09\%$\uparrow$ \\
\rowcolor{experimentours}
\textbf{Rubrics on Trial} & \textbf{79.10} & \textbf{2.66\%$\uparrow$} & 54.12 & 7.00\%$\uparrow$ & \textbf{43.95} & \textbf{9.38\%$\uparrow$} & \textbf{76.14} & \textbf{4.24\%$\uparrow$} & \textbf{74.83} & \textbf{3.70\%$\uparrow$} \\
\bottomrule
}
\setlength{\tabcolsep}{0pt}
\setbox0=\hbox{\begin{tabular}{l*{10}{c}}\experimentrows\end{tabular}}
\dimen0=\dimexpr\linewidth-\wd0\relax
\divide\dimen0 by 22
\setlength{\tabcolsep}{\dimen0}
\begin{tabular}{l*{10}{c}}\experimentrows\end{tabular}
\par\vspace{3pt}\raggedright\scriptsize
\textit{Note:} $\Delta$ indicates the relative improvement over the Original policy in percentage. The best results in each column are \textbf{bolded}.
\end{table}

\endgroup

\subsection{Evolution Sensitivity and Cost}
\label{subsec:evolution_sensitivity}

In this section, we examine how the evolution budget affects rubric quality.
Figure~\ref{fig:evolution_sensitivity} presents the results over one to fifteen rounds, with two key observations.
\textbf{\ding{182} Rubric evolution enhances rubric quality.}
Preference accuracy increases across all three benchmarks during the early rounds.
This supports successive proposal and comparison as an effective way to improve LLM-generated rubrics.
\textbf{\ding{183} Rubric evolution exhibits empirical convergence in performance.}
In later rounds, preference accuracy largely stabilizes, while the average number of criteria per rubric continues to grow at a slower pace.
This suggests that a moderate number of evolution rounds is sufficient to obtain high-quality rubrics.
We also measure cumulative output tokens and active request time per query during rubric evolution (Figure~\ref{fig:evolution_cost}).
Both costs continue to grow after accuracy largely stabilizes, supporting a bounded budget to balance rubric quality and generation cost.

\begingroup
\setlength{\intextsep}{0pt}
\begin{figure}[!ht]
\captionsetup{skip=4pt}
\centering
\begin{minipage}[t]{0.49\linewidth}
\vspace{0pt}
\centering
\includegraphics[width=\linewidth]{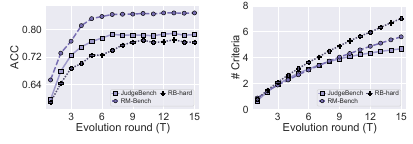}
\caption{Varying evolution round results.}
\label{fig:evolution_sensitivity}
\end{minipage}\hfill
\begin{minipage}[t]{0.49\linewidth}
\vspace{0pt}
\centering
\includegraphics[width=\linewidth]{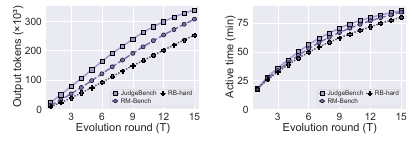}
\caption{Varying evolution round costs.}
\label{fig:evolution_cost}
\end{minipage}
\end{figure}
\endgroup
\Needspace{6\baselineskip}
\section{Conclusion}
\label{sec:conclusion}

In this paper, we identify two key challenges in rubric comparison: reflecting downstream task performance and discouraging unnecessary criteria. To address these challenges, we introduce Rubrics on Trial, a framework that compares rubrics using synthetic response pairs and alternative responses. Theoretically, the favored rubric yields a better policy after RLHF.
Experiments show that the proposed method achieves state-of-the-art performance in rubric evolution, the evolution enhances the quality of generated rubrics, and the resulting rubrics improve downstream RLHF performance.

\textbf{Limitations \& Future Work.}
Two limitations of this work warrant future investigation:
(1) Rubrics on Trial evolves a rubric for each query separately, so the evolution cost grows with the number of rounds; distilling evolved rubrics into a trained rubric generator may reduce this cost.
(2) Rubric comparison relies on an LLM comparator, which may misjudge responses whose quality the comparator cannot verify, such as complex code; incorporating verifiable signals such as code execution remains a valuable direction.

\small
\bibliography{iclr2027_conference}
\bibliographystyle{icml2026}
\normalsize

\clearpage
\appendix
\section{Related Work}
\label{app:related_work}

This work studies rubric-based evaluation, which assesses LLM response quality using explicit criteria.
Rubrics support calibrated evaluation~\citep{hashemi2024llmrubric,rao2026autorubric} and reward construction for policy optimization~\citep{gunjal2026rubrics}.
Beyond manually designed rubrics, rubric construction methods broadly fall into two categories: \textbf{trained rubric generators}, which are models trained to generate rubrics, and \textbf{LLM-based rubric generation}, which prompts fixed LLMs to construct rubrics.
Rubrics on Trial belongs to the second category and studies query-only rubric evolution.

\subsection{Trained Rubric Generators}
\label{app:trained_rubric_generators}

OpenRubrics~\citep{liu2025openrubrics} constructs rubric training data through contrastive generation and preference-consistency filtering.
RubricARM~\citep{xu2026rubricarm} and RubricARROW~\citep{jiang2026rubricarrow} alternate generator and judge training to improve preference prediction.
For domain-specific generation, \citet{lv2026queryspecific} train query-specific rubric generators from human preferences over DeepResearch reports.
EvoRubric~\citep{guan2026evorubric} further couples rubric generation and response generation within a shared policy trained by reinforcement learning.
Although often effective in practice, these methods require training a rubric generator.
OpenRubrics also requires curated rubric data, while the DeepResearch generator requires human preference annotations.
Rubrics on Trial instead uses fixed LLMs to improve rubrics for each query through proposal and comparison, avoiding this training cost.

\subsection{Training-Free Rubric Generation}
\label{app:automatic_rubric_generation}

To avoid the cost of training a rubric generator, a prevalent strategy is to prompt an LLM to generate a rubric in a single pass.
TICK~\citep{cook2024ticking} and RocketEval~\citep{wei2025rocketeval} generate query-specific checklists for response evaluation.
RubricRAG~\citep{dhole2026rubricrag} guides generation with rubrics retrieved from related queries.
A more recent line focuses on rubric evolution.
Chasing the Tail~\citep{zhang2026chasing} refines rubrics by identifying differences between high-quality responses.
RRD~\citep{shen2026rethinking} repeatedly breaks broad criteria into more specific ones and filters out criteria that overlap or favor responses from weaker models.
Qworld~\citep{gao2026qworld} expands each query into scenarios, perspectives, and detailed criteria over multiple rounds.
However, rubric comparison should reflect downstream policy performance without repeated RLHF training and discourage unnecessary criteria~\citep{mahmoud2026rewardhacking,shen2026rethinking}.
Rubrics on Trial addresses both by using synthetic response comparisons as a proxy for downstream evaluation and alternative responses to help identify unnecessary criteria.

\subsection{Reinforcement Learning for LLMs}
\label{app:llm_reinforcement_learning}

Reinforcement learning improves LLM behavior by optimizing rewards derived from human preferences~\citep{ouyang2022training} or AI feedback~\citep{bai2022constitutional}.
DeepSeekMath~\citep{shao2024deepseekmath} and DeepSeek-R1~\citep{guo2025deepseekr1} improve reasoning, with the latter using verifiable rewards for automatically checkable tasks.
For learned rewards, recent work addresses biased feedback~\citep{wangunbiased,wang2026causalrm}, label noise~\citep{pan2026optimal}, and reward uncertainty~\citep{pan2026uncertainty} to improve the signals used for policy optimization.
Related LLM applications include multimodal reasoning~\citep{wang2025video,wang2026museagent} and GUI interaction~\citep{liu2026memgui,liu2025llmpowered}, with rubrics also used to evaluate agent behavior~\citep{chen2026causalgame}.

For open-ended tasks, however, a single correctness check is often insufficient to assess response quality.
Rubric-based RL addresses this limitation by using explicit criteria to construct reward signals.
Rubrics as Rewards~\citep{gunjal2026rubrics} and Rubric Anchors~\citep{huang2025rubricanchors} use task-specific rubrics to guide policy optimization beyond verifiable domains.
Reinforcement Learning from Checklist Feedback~\citep{viswanathan2025checklists} extracts instruction-specific checklists and uses item-level judgments for preference optimization.
DR Tulu~\citep{shao2026drtulu} further evolves rubrics alongside the policy, incorporating newly discovered information into the training feedback.
These methods, however, depend on the quality of the criteria used to define rewards~\citep{mahmoud2026rewardhacking}.
Rubrics on Trial evolves these criteria before policy training, with downstream RL experiments testing whether better rubrics improve policy performance.

\section{Theoretical Justification}
\label{app:theoretical_justification}

In this section, we prove Theorem~\ref{thm:rank_preservation}, extend the analysis to policies whose responses may still violate rubric criteria after RLHF (Theorem~\ref{thm:incomplete_rank_preservation}), and connect these results to ADD comparisons (Corollary~\ref{cor:add_rank_preservation}).
Throughout, we fix the query $x$ and use the definitions of $\mu(r)$ and $\hat{\mu}(r)$ in Eqs.~\ref{eq:post_rl_quality} and~\ref{eq:rubric_satisfying_quality}.
The initial policy $\pi(\cdot\mid x)$ is defined over a countable response space $\mathcal Y$, and the quality function $g(x,y)\in[0,M]$ is independent of the rubric used as the training reward.
Let $S_r=\{y\in\mathcal Y:y\models r\}$ contain responses satisfying every criterion in $r$, and assume $\pi(S_r\mid x)>0$.
We write $\mathbb E_\pi$ for expectation under $\pi(\cdot\mid x)$ and $\rho(r)=\mathbb P_{y\sim\pi_r(\cdot\mid x)}(y\models r)$ for the trained satisfaction probability.

All optimizations below range over distributions $q$ absolutely continuous with respect to $\pi(\cdot\mid x)$.
The KL order is $\KL(q\Vert\pi)$: reversing it would give infinite divergence under complete satisfaction whenever $\pi(S_r\mid x)<1$.

\subsection{Proof of Theorem~\ref{thm:rank_preservation}}
\label{app:proof_rank_preservation}

\begin{rankpreservationrestate}
Assume that rubric satisfaction has positive probability under the initial policy $\pi$, so that $\hat{\mu}$ is well-defined. For each $r\in\{r_t,\hat r_{t+1}\}$, suppose the post-RLHF policy minimizes KL divergence from $\pi$ among policies that fully satisfy its reward rubric:
\[
\pi_r(\cdot\mid x)
=\operatorname*{arg\,min}_{q:\,\mathbb P_{y\sim q}(y\models r)=1}
\KL\!\left(q\,\Vert\,\pi(\cdot\mid x)\right).
\]
Then $\pi_r(\cdot\mid x)=\pi(\cdot\mid x,y\models r)$ and $\mu(r)=\hat{\mu}(r)$. Consequently,
\[
\hat{\mu}(r_t)>\hat{\mu}(\hat r_{t+1})
\quad\Longleftrightarrow\quad
\mu(r_t)>\mu(\hat r_{t+1}).
\]
\end{rankpreservationrestate}

\begin{proof}
By conditional probability,
\begin{equation}
\label{eq:conditional_probability_normalization}
\pi(y\mid x,S_r)=\frac{\pi(y\mid x)}{\pi(S_r\mid x)}
\quad\text{for }y\in S_r,
\end{equation}
and it is zero outside $S_r$. The denominator normalizes the satisfying
responses to total probability one. For any feasible $q$, $q(S_r)=1$,
so substituting $\pi(y\mid x)=\pi(y\mid x,S_r)\pi(S_r\mid x)$ gives
\begin{align}
\KL\!\left(q\,\Vert\,\pi(\cdot\mid x)\right)
&=\sum_{y\in S_r}q(y)
  \log\frac{q(y)}{\pi(y\mid x,S_r)\pi(S_r\mid x)}
  \nonumber\\
&=\KL\!\left(q\,\Vert\,\pi(\cdot\mid x,S_r)\right)
  -\log\pi(S_r\mid x).
\label{eq:complete_kl_decomposition}
\end{align}
The first term is nonnegative and equals zero exactly when
$q=\pi(\cdot\mid x,S_r)$. This distribution is feasible, so it is the
unique minimizer and hence equals $\pi_r$. Therefore,
\[
\mu(r)
=\mathbb E_{y\sim\pi_r(\cdot\mid x)}[g(x,y)]
=\mathbb E_{y\sim\pi(\cdot\mid x,S_r)}[g(x,y)]
=\hat\mu(r).
\]
Applying this identity to both rubrics proves the claim.
\end{proof}

\subsection{Extension to Incomplete Rubric Satisfaction}
\label{app:imperfect_rubric_satisfaction}

We next extend Theorem~\ref{thm:rank_preservation} to the more general setting in which policies may not fully satisfy their reward rubrics after RLHF. Specifically, we relax complete satisfaction to a shared lower bound $\alpha\in(0,1]$ on the satisfaction probability, requiring $\rho(r)\ge\alpha$ for both rubrics.

\begin{theorem}[Rank preservation under incomplete satisfaction]
\label{thm:incomplete_rank_preservation}
For each $r\in\{r_t,\hat r_{t+1}\}$, suppose
\begin{equation}
\label{eq:incomplete_projection}
\pi_r(\cdot\mid x)
=\operatorname*{arg\,min}_{q:\,q(S_r)\ge\alpha}
\KL\!\left(q\,\Vert\,\pi(\cdot\mid x)\right).
\end{equation}
Assume
\begin{equation}
\label{eq:incomplete_rank_assumptions}
\pi(S_{\hat r_{t+1}}\mid x)\le\pi(S_{r_t}\mid x),
\qquad
\hat\mu(r_t)\ge\mathbb E_\pi[g(x,y)].
\end{equation}
If $\alpha>\pi(S_{\hat r_{t+1}}\mid x)$, then
\begin{equation}
\label{eq:incomplete_rank_preservation}
\hat\mu(\hat r_{t+1})>\hat\mu(r_t)
\quad\Longrightarrow\quad
\mu(\hat r_{t+1})>\mu(r_t).
\end{equation}
If $\alpha\le\pi(S_{\hat r_{t+1}}\mid x)$, both trained policies equal
$\pi$ and their qualities tie.
\end{theorem}

The two conditions in Eq.~\ref{eq:incomplete_rank_assumptions} say that
the candidate is no more likely to be satisfied under $\pi$, and that
responses satisfying the current rubric have at least the initial
policy's average quality.

\begin{proof}
First fix $r$ with $0<\pi(S_r\mid x)<1$. Splitting the KL divergence
over $S_r$ and its complement gives
\begin{equation}
\label{eq:incomplete_kl_decomposition}
\begin{aligned}
\KL\!\left(q\,\Vert\,\pi(\cdot\mid x)\right)
={}&q(S_r)\log\frac{q(S_r)}{\pi(S_r\mid x)}
 +[1-q(S_r)]\log\frac{1-q(S_r)}{1-\pi(S_r\mid x)}\\
&+q(S_r)\,\KL\!\left(q(\cdot\mid S_r)\,\Vert\,
                    \pi(\cdot\mid x,S_r)\right)\\
&+[1-q(S_r)]\,\KL\!\left(q(\cdot\mid S_r^c)\,\Vert\,
                        \pi(\cdot\mid x,S_r^c)\right).
\end{aligned}
\end{equation}
Terms with zero mass contribute zero. For a fixed $q(S_r)$, the last
two terms are minimized by preserving $\pi$'s conditional distributions.
The first line is a strictly convex function of $q(S_r)$, minimized at
$q(S_r)=\pi(S_r\mid x)$. Constraining $q(S_r)\ge\alpha$ therefore gives
the unique solution
\begin{equation}
\label{eq:incomplete_conditional_policy}
\begin{aligned}
\rho(r)&=\max\{\pi(S_r\mid x),\alpha\},\\
\pi_r(\cdot\mid x)
&=\rho(r)\pi(\cdot\mid x,S_r)
  +[1-\rho(r)]\pi(\cdot\mid x,S_r^c).
\end{aligned}
\end{equation}
If $\pi(S_r\mid x)=1$, the initial policy is already feasible and
uniquely minimizes KL, so $\mu(r)=\hat\mu(r)=\mathbb E_\pi[g(x,y)]$.

For $\pi(S_r\mid x)<1$, the initial expectation decomposes as
\[
\mathbb E_\pi[g(x,y)]
=\pi(S_r\mid x)\hat\mu(r)
 +[1-\pi(S_r\mid x)]\mathbb E_\pi[g(x,y)\mid y\not\models r].
\]
Taking expectations under Eq.~\ref{eq:incomplete_conditional_policy}
and subtracting this identity yields
\begin{align}
\mu(r)-\mathbb E_\pi[g(x,y)]
&=[\rho(r)-\pi(S_r\mid x)]
  \bigl[\hat\mu(r)-\mathbb E_\pi[g(x,y)\mid y\not\models r]\bigr]
\label{eq:exact_quality_gain}\\
&=\frac{\rho(r)-\pi(S_r\mid x)}{1-\pi(S_r\mid x)}
  \bigl[\hat\mu(r)-\mathbb E_\pi[g(x,y)]\bigr].
\label{eq:quality_from_surrogate}
\end{align}
The first equality expresses the quality gain as the increase in
satisfaction probability times the quality gap between satisfying and
violating responses.

The coefficient in Eq.~\ref{eq:quality_from_surrogate} is nonnegative
and nonincreasing in $\pi(S_r\mid x)$: it is zero above $\alpha$, and
below $\alpha$ its derivative is
$(\alpha-1)/(1-\pi(S_r\mid x))^2\le0$.
Suppose $\alpha>\pi(S_{\hat r_{t+1}}\mid x)$ and
$\pi(S_{r_t}\mid x)<1$. By Eq.~\ref{eq:incomplete_rank_assumptions},
the candidate's coefficient is at least the current rubric's.
Subtracting Eq.~\ref{eq:quality_from_surrogate} for the two rubrics,
using $\hat\mu(r_t)-\mathbb E_\pi[g(x,y)]\ge0$, gives
\begin{equation}
\label{eq:incomplete_rank_bound}
\mu(\hat r_{t+1})-\mu(r_t)
\ge
\frac{\alpha-\pi(S_{\hat r_{t+1}}\mid x)}
     {1-\pi(S_{\hat r_{t+1}}\mid x)}
\bigl[\hat\mu(\hat r_{t+1})-\hat\mu(r_t)\bigr].
\end{equation}
The right-hand side is strictly positive when the surrogate ranks the
candidate higher. If $\pi(S_{r_t}\mid x)=1$, the same bound follows
directly from $\mu(r_t)=\hat\mu(r_t)=\mathbb E_\pi[g(x,y)]$.
Finally, if $\alpha\le\pi(S_{\hat r_{t+1}}\mid x)$, the initial policy
is feasible for both rubrics and uniquely minimizes both KL objectives.
\end{proof}

\subsection{Rank Preservation for ADD}
\label{app:add_rank_preservation}

For $\hat r_{t+1}=r_t\cup\{c\}$, define
\begin{equation}
\label{eq:add_contrast}
\begin{aligned}
p_c&=\mathbb P_\pi(y\models c\mid y\models r_t),\\
\Delta_c&=\hat\mu(\hat r_{t+1})
 -\mathbb E_\pi[g(x,y)\mid y\models r_t,\ y\not\models c].
\end{aligned}
\end{equation}
Assume $0<p_c<1$, so both satisfying and violating subsets within $S_{r_t}$
have positive probability.

\begin{corollary}[Rank preservation for ADD]
\label{cor:add_rank_preservation}
For an ADD proposal,
\begin{equation}
\label{eq:add_surrogate_identity}
\hat\mu(\hat r_{t+1})-\hat\mu(r_t)=(1-p_c)\Delta_c.
\end{equation}
Under the complete-satisfaction model of Theorem~\ref{thm:rank_preservation},
\begin{equation}
\label{eq:add_rank_preservation}
\Delta_c>0
\ \Longrightarrow\
\hat\mu(\hat r_{t+1})>\hat\mu(r_t)
\ \Longrightarrow\
\mu(\hat r_{t+1})>\mu(r_t).
\end{equation}
The same implication holds under Eq.~\ref{eq:incomplete_projection}
if $\hat\mu(r_t)\ge\mathbb E_\pi[g(x,y)]$ and
$\alpha>\pi(S_{\hat r_{t+1}}\mid x)$.
If $\alpha\le\pi(S_{\hat r_{t+1}}\mid x)$, both policies
equal $\pi$.
\end{corollary}

\begin{proof}
Partitioning $S_{r_t}$ by satisfaction of $c$ gives
\[
\hat\mu(r_t)
=p_c\hat\mu(\hat r_{t+1})
 +(1-p_c)\mathbb E_\pi[g(x,y)\mid y\models r_t,\ y\not\models c].
\]
Subtracting this expression from $\hat\mu(\hat r_{t+1})$ proves
Eq.~\ref{eq:add_surrogate_identity}. Moreover,
\[
S_{\hat r_{t+1}}\subseteq S_{r_t},
\qquad
\pi(S_{\hat r_{t+1}}\mid x)=\pi(S_{r_t}\mid x)p_c
\le\pi(S_{r_t}\mid x).
\]
Thus Theorem~\ref{thm:rank_preservation} gives the complete-satisfaction
claim, and Theorem~\ref{thm:incomplete_rank_preservation} gives the
incomplete-satisfaction claim.
\end{proof}

For the empty rubric, $S_{r_0}=\mathcal Y$ and
$\hat\mu(r_0)=\mathbb E_\pi[g(x,y)]$.
With fixed $\pi$, $g$, and $\alpha$, a sequence of ADD steps with
$\Delta_c>0$ preserves $\hat\mu(r_t)\ge\mathbb E_\pi[g(x,y)]$ by
Eq.~\ref{eq:add_surrogate_identity}. Its projected policy quality is
therefore nondecreasing, with strict improvement whenever the candidate
constraint requires an increase in satisfaction probability.

\paragraph{Connection to the implemented comparison.}
The ADD pair has $a^+\models\hat r_{t+1}$ and
$a^-\models r_t$ but $a^-\not\models c$. If its marginals satisfy
\[
a^+\sim\pi(\cdot\mid x,S_{\hat r_{t+1}}),
\qquad
a^-\sim\pi(\cdot\mid x,\ y\models r_t,\ y\not\models c),
\]
then $\mathbb E[g(x,a^+)-g(x,a^-)]=\Delta_c$.

\section{More Experimental Results}
\label{app:more_experimental_results}

\Needspace{27\baselineskip}
\subsection{Generalization across Verifier LLMs}
\label{app:verifier_generalization}

\begin{wraptable}{r}{0.5\textwidth}
\vspace{-6pt}
\centering
\captionsetup{font=small,aboveskip=0pt,belowskip=4pt}
\caption{Performance across verifier LLMs.}
\label{tab:verifier_generalization}
\renewcommand{\arraystretch}{1.03}
\fontsize{7.5}{8.5}\selectfont
\def\experimentrows{
\toprule
\multirow{2}{*}{\textbf{Method}} & \multicolumn{2}{c}{\textbf{JudgeBench}} & \multicolumn{2}{c}{\textbf{RM-Bench}} & \multicolumn{2}{c}{\textbf{Chat-Hard}} \\
\cmidrule(lr){2-3}\cmidrule(lr){4-5}\cmidrule(lr){6-7}
 & ACC & $\Delta$ & ACC & $\Delta$ & ACC & $\Delta$ \\
\midrule
\rowcolor{experimentgray}\multicolumn{7}{l}{\textbf{\textit{Verifier: GPT-4.1-mini}}} \\
\rowcolor{white}
Direct & 67.34 & - & 77.72 & - & 73.25 & - \\
\rowcolor{experimentours}
\textbf{Ours} & \textbf{78.79} & 17.01\%$\uparrow$ & \textbf{84.75} & 9.05\%$\uparrow$ & \textbf{76.97} & 5.08\%$\uparrow$ \\
\midrule
\rowcolor{experimentgray}\multicolumn{7}{l}{\textbf{\textit{Verifier: Gemini-3.1-Flash-Lite}}} \\
\rowcolor{white}
Direct & 77.74 & - & 82.32 & - & 75.11 & - \\
\rowcolor{experimentours}
\textbf{Ours} & \textbf{80.81} & 3.94\%$\uparrow$ & \textbf{85.84} & 4.28\%$\uparrow$ & \textbf{78.84} & 4.96\%$\uparrow$ \\
\midrule
\rowcolor{experimentgray}\multicolumn{7}{l}{\textbf{\textit{Verifier: GPT-4.1}}} \\
\rowcolor{white}
Direct & 71.13 & - & 78.96 & - & 72.48 & - \\
\rowcolor{experimentours}
\textbf{Ours} & \textbf{79.52} & 11.79\%$\uparrow$ & \textbf{83.72} & 6.03\%$\uparrow$ & \textbf{76.54} & 5.60\%$\uparrow$ \\
\midrule
\rowcolor{experimentgray}\multicolumn{7}{l}{\textbf{\textit{Verifier: Qwen3-Max}}} \\
\rowcolor{white}
Direct & 72.02 & - & 80.53 & - & 74.12 & - \\
\rowcolor{experimentours}
\textbf{Ours} & \textbf{79.60} & 10.53\%$\uparrow$ & \textbf{84.95} & 5.49\%$\uparrow$ & \textbf{78.73} & 6.21\%$\uparrow$ \\
\midrule
\rowcolor{experimentgray}\multicolumn{7}{l}{\textbf{\textit{Verifier: GLM-5.1}}} \\
\rowcolor{white}
Direct & 79.52 & - & 84.37 & - & 75.33 & - \\
\rowcolor{experimentours}
\textbf{Ours} & \textbf{82.18} & 3.35\%$\uparrow$ & \textbf{87.00} & 3.12\%$\uparrow$ & \textbf{78.84} & 4.66\%$\uparrow$ \\
\bottomrule
}
\setlength{\tabcolsep}{0pt}
\setbox0=\hbox{\begin{tabular}{l*{6}{c}}\experimentrows\end{tabular}}
\dimen0=\dimexpr\linewidth-\wd0\relax
\divide\dimen0 by 14
\setlength{\tabcolsep}{\dimen0}
\begin{tabular}{l*{6}{c}}\experimentrows\end{tabular}
\par\vspace{3pt}\raggedright\scriptsize
\textit{Note:} ACC is preference accuracy (\%). $\Delta$ indicates the relative improvement over Direct for the same verifier in percentage. The best ACC for each verifier is \textbf{bolded}.
\vspace{-6pt}
\end{wraptable}

In this section, we examine whether the gains in rubric-based evaluation generalize across verifier LLMs.
To complement the study of generation LLMs in Section~\ref{subsec:llm_generalization}, we keep the generated rubrics fixed and vary only the final verifier.
Each verifier checks criterion satisfaction and uses the resulting rubric scores to rank the candidate responses.
We compare Rubrics on Trial with Direct on JudgeBench, RM-Bench, and Chat-Hard, using GPT-4.1-mini from the main experiments as a reference and four alternative verifiers from different model families.

Table~\ref{tab:verifier_generalization} presents the results.
\textbf{Rubric quality improves across verifiers.}
Rubrics on Trial outperforms Direct in all fifteen verifier--benchmark combinations, with relative gains ranging from $3.12\%$ to $11.79\%$.
The gains remain positive on all three benchmarks when GPT-4.1-mini is replaced by any of the four alternative verifiers, supporting improved rubric quality under different models that assess criterion satisfaction.
Together with Section~\ref{subsec:llm_generalization}, these findings support the generalizability of Rubrics on Trial across the LLMs used for both rubric generation and evaluation.

\subsection{Case Study of Alternative Responses}
\label{app:alternative_case}

In this section, we present the complete case discussed in Section~\ref{subsec:alternative_comparison}, as shown in Table~\ref{tab:alternative_case}.
The additional criterion $c$ requires two separate reasoning steps, which are not essential for satisfying the user query.
The comparison between $a^+$ and $a^-$ favors $c$: editing $a^+$ to violate $c$ removes the entire justification, so the comparator prefers $a^+$ for justifying the answer rather than for using two separate steps.
The comparison between $b^+$ and $b^-$ reveals that $c$ is unnecessary: the alternative response $b^-$ already justifies the answer in a single step, and editing $b^-$ to satisfy $c$ splits the same justification into two steps without improving response quality.
The comparator therefore finds $b^+$ and $b^-$ equally good, so $\hat d_t=0$ in Eq.~\ref{eq:rubric_comparison_signal} and $c$ is rejected.
In contrast, for a necessary criterion such as $c_1$, any $b^-$ that violates $c_1$ does not select the correct option (C), so editing $b^-$ to select (C) improves response quality and yields $b^+\succ b^-$.

\begin{table}[H]
\caption{An illustrative case of rubric comparison with an unnecessary criterion.}
\label{tab:alternative_case}
\centering
\footnotesize
\renewcommand{\arraystretch}{1.2}
\setlength{\tabcolsep}{4pt}
\setlength{\caselabelwidth}{2.3cm}
\setlength{\casecheckwidth}{0.75cm}
\setlength{\casetextwidth}{\dimexpr\linewidth-\caselabelwidth-2\casecheckwidth-8\tabcolsep\relax}
\setlength{\casespanwidth}{\dimexpr\casetextwidth+2\casecheckwidth+4\tabcolsep\relax}
\begin{tabular}{>{\raggedright\arraybackslash}p{\caselabelwidth}p{\casetextwidth}>{\centering\arraybackslash}p{\casecheckwidth}>{\centering\arraybackslash}p{\casecheckwidth}}
\toprule
\textbf{Query $x$} & \multicolumn{3}{p{\casespanwidth}}{\casequery{Which of the following is a prime number? (A) 21 (B) 27 (C) 29 (D) 33}} \\
\textbf{Current rubric} & \multicolumn{3}{p{\casespanwidth}}{$r_t=\{c_1\}$, where $c_1$: Selects (C) 29.} \\
\textbf{Additional criterion} & \multicolumn{3}{p{\casespanwidth}}{$c$: Checks the options in at least two separate reasoning steps before giving the letter.} \\
\midrule
 & & \multicolumn{2}{c}{\textbf{Verifier}} \\
\cmidrule(lr){3-4}
\textbf{Response} & \textbf{Text} & $c_1$ & $c$ \\
\midrule
\multicolumn{4}{l}{\textit{Comparison between $a^+$ and $a^-$}} \\
$a^+$ & \casesatisfy{$21=3\times7$, $27=3\times9$, and $33=3\times11$, so they are composite. 29 has no divisor other than 1 and itself, so it is prime.} Answer: (C). & \Checkmark & \Checkmark \\
$a^-$ & Answer: (C). & \Checkmark & \XSolidBrush \\
Comparator & \multicolumn{3}{p{\casespanwidth}}{\casejudge{Both responses select (C), but only $a^+$ gives a correct justification. $a^+$ is better.} $\Rightarrow$ $a^+\succ a^-$} \\
\addlinespace[2pt]
\hdashline
\addlinespace[1pt]
\multicolumn{4}{l}{\textit{Comparison between $b^+$ and $b^-$}} \\
$b^-$ & 29 is the only option not divisible by 3 (21, 27, and 33 all are), so the answer is (C). & \Checkmark & \XSolidBrush \\
$b^+$ & \casesatisfy{21, 27, and 33 are all divisible by 3, so they are not prime. 29 is not divisible by 3, so it is the only remaining option.} Answer: (C). & \Checkmark & \Checkmark \\
Comparator & \multicolumn{3}{p{\casespanwidth}}{\casejudge{Both responses select (C) with a correct justification. Neither has a real advantage.} $\Rightarrow$ tie} \\
\midrule
\textbf{Decision} & \multicolumn{3}{p{\casespanwidth}}{$a^+\succ a^-$ but $b^+\not\succ b^-$, so $\hat d_t=0$ and $r_{t+1}=r_t$; the unnecessary criterion $c$ is rejected.} \\
\bottomrule
\end{tabular}
\par\vspace{3pt}\raggedright\scriptsize
\textit{Note:} \casequery{The query is in blue}, \casesatisfy{reasoning steps that satisfy $c$ are in green}, and \casejudge{comparator reasons are in orange}. Comparator reasons are abridged, with the response labels restored for readability. \Checkmark\ and \XSolidBrush\ denote the verifier results for $c_1$ and $c$.
\end{table}

\subsection{Robustness to Response Style}
\label{app:response_style}

To examine Challenge~\ding{183}, we focus on one category of unnecessary criteria: stylistic requirements that are not essential for satisfying the user query.
On RM-Bench~\citep{liu2024rmbench}, we fix the chosen response to its concise version and vary the rejected response among Concise, Plain, and Markdown.
The latter two are hard settings: rejected responses gain length or formatting advantages despite inferior content.
We compare Rubrics on Trial, its variant without $(b^+,b^-)$, and Direct using the same examples and verifier to assess robustness to these advantages.

\begin{figure}[H]
    \centering
    \captionsetup[subfigure]{font=small,skip=2pt}
    \begin{subfigure}[t]{0.248\linewidth}
        \centering
        \includegraphics[width=\linewidth]{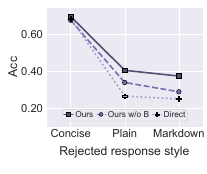}
        \caption{Chat.}
    \end{subfigure}\hfill
    \begin{subfigure}[t]{0.248\linewidth}
        \centering
        \includegraphics[width=\linewidth]{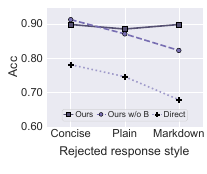}
        \caption{Math.}
    \end{subfigure}\hfill
    \begin{subfigure}[t]{0.248\linewidth}
        \centering
        \includegraphics[width=\linewidth]{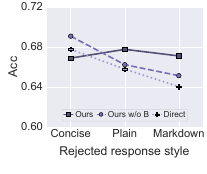}
        \caption{Code.}
    \end{subfigure}\hfill
    \begin{subfigure}[t]{0.248\linewidth}
        \centering
        \includegraphics[width=\linewidth]{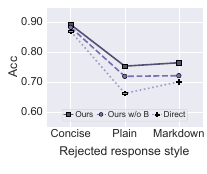}
        \caption{Safety.}
    \end{subfigure}
    \caption{Robustness to response style on RM-Bench. Ours and Ours w/o B denote Rubrics on Trial and its variant without $(b^+,b^-)$. Plain and Markdown denote detailed plain-text and Markdown responses, respectively.}
    \label{fig:rmbench_response_style}
\end{figure}

Figure~\ref{fig:rmbench_response_style} presents two key observations.
\textbf{\ding{182} Style advantages can distort rubric-based preferences.}
When the rejected response changes from Concise to Plain or Markdown, preference accuracy generally falls, most sharply in Chat and Safety.
This highlights the need to discourage unnecessary stylistic criteria.
\textbf{\ding{183} Alternative responses mitigate style interference.}
Despite this decline, Rubrics on Trial is less sensitive and outperforms both Direct and the variant without $(b^+,b^-)$ under Plain and Markdown across all domains.
These results suggest that comparison with alternative responses helps discourage unnecessary stylistic criteria, thereby improving rubric quality.

\subsection{Out-of-Domain Generalization}
\label{app:rl_generalization}

We further examine whether the downstream policy improvements generalize across domains.
We evaluate the WildChat-trained policies from Section~\ref{subsec:downstream_rl} on HealthBench-Hard~\citep{arora2025healthbench}, comprising 1{,}000 healthcare conversations, using GPT-4.1 and physician-written criteria.
Table~\ref{tab:healthbench_generalization} presents the results with two key observations.
\textbf{\ding{182} Policy improvements generalize to a specialized domain.}
Policies trained with Rubrics on Trial outperform their original counterparts across all three models.
This suggests that improving rubric quality can support policy gains that generalize from general-purpose tasks to a specialized domain.
\textbf{\ding{183}~The advantage over other rubric rewards persists under domain shift.}
Rubrics on Trial also achieves the highest average score among the compared methods for all three policies.
Together with the main RLHF results, these findings support the effectiveness of our improved rubrics as rewards for policy learning with benefits across domains.

\begin{table}[H]
\caption{Out-of-domain generalization of RLHF-trained policies on HealthBench-Hard.}
\label{tab:healthbench_generalization}
\centering
\renewcommand{\arraystretch}{1.00}
\fontsize{7.8}{10}\selectfont
\def\experimentrows{
\toprule
\multirow{2}{*}{\textbf{Method}} & \multicolumn{2}{c}{\textbf{Accuracy}} & \multicolumn{2}{c}{\shortstack{\textbf{Communication}\\\textbf{Quality}}} & \multicolumn{2}{c}{\textbf{Completeness}} & \multicolumn{2}{c}{\shortstack{\textbf{Context}\\\textbf{Awareness}}} & \multicolumn{2}{c}{\shortstack{\textbf{Instruction}\\\textbf{Following}}} & \multicolumn{2}{c}{\textbf{Average}} \\
\cmidrule(lr){2-3}\cmidrule(lr){4-5}\cmidrule(lr){6-7}\cmidrule(lr){8-9}\cmidrule(lr){10-11}\cmidrule(lr){12-13}
 & Value & $\Delta$ & Value & $\Delta$ & Value & $\Delta$ & Value & $\Delta$ & Value & $\Delta$ & Value & $\Delta$ \\
\midrule
\rowcolor{experimentgray}\multicolumn{13}{l}{\textbf{\textit{Policy Model: Qwen2.5-3B-Instruct}}} \\
\rowcolor{white}
Original & 19.62 & - & 46.12 & - & 10.44 & - & 13.00 & - & 36.38 & - & 25.11 & - \\
\rowcolor{white}
Direct & 20.97 & 6.88\%$\uparrow$ & 45.44 & 1.47\%$\downarrow$ & 12.16 & 16.48\%$\uparrow$ & \textbf{14.48} & \textbf{11.38\%$\uparrow$} & 36.30 & 0.22\%$\downarrow$ & 25.87 & 3.02\%$\uparrow$ \\
\rowcolor{white}
TICK & 20.01 & 1.99\%$\uparrow$ & 46.82 & 1.52\%$\uparrow$ & 11.50 & 10.15\%$\uparrow$ & 13.12 & 0.92\%$\uparrow$ & 36.54 & 0.44\%$\uparrow$ & 25.60 & 1.94\%$\uparrow$ \\
\rowcolor{white}
RocketEval & 20.41 & 4.03\%$\uparrow$ & 48.46 & 5.07\%$\uparrow$ & 12.93 & 23.85\%$\uparrow$ & 13.62 & 4.77\%$\uparrow$ & 38.28 & 5.22\%$\uparrow$ & 26.74 & 6.48\%$\uparrow$ \\
\rowcolor{white}
ChaseTail & 20.95 & 6.78\%$\uparrow$ & 48.39 & 4.92\%$\uparrow$ & \textbf{13.98} & \textbf{33.91\%$\uparrow$} & 13.47 & 3.62\%$\uparrow$ & 37.80 & 3.90\%$\uparrow$ & 26.92 & 7.19\%$\uparrow$ \\
\rowcolor{experimentours}
\textbf{Rubrics on Trial} & \textbf{21.04} & \textbf{7.24\%$\uparrow$} & \textbf{49.96} & \textbf{8.33\%$\uparrow$} & 13.96 & 33.72\%$\uparrow$ & 14.13 & 8.69\%$\uparrow$ & \textbf{38.59} & \textbf{6.07\%$\uparrow$} & \textbf{27.54} & \textbf{9.65\%$\uparrow$} \\
\midrule
\rowcolor{experimentgray}\multicolumn{13}{l}{\textbf{\textit{Policy Model: Qwen2.5-7B-Instruct}}} \\
\rowcolor{white}
Original & 22.39 & - & 51.12 & - & 13.88 & - & 15.69 & - & 41.99 & - & 29.01 & - \\
\rowcolor{white}
Direct & 22.74 & 1.56\%$\uparrow$ & 52.55 & 2.80\%$\uparrow$ & 15.36 & 10.66\%$\uparrow$ & 15.89 & 1.27\%$\uparrow$ & 42.35 & 0.86\%$\uparrow$ & 29.78 & 2.63\%$\uparrow$ \\
\rowcolor{white}
TICK & 22.89 & 2.23\%$\uparrow$ & 50.15 & 1.90\%$\downarrow$ & 15.91 & 14.63\%$\uparrow$ & 16.16 & 3.00\%$\uparrow$ & 38.16 & 9.12\%$\downarrow$ & 28.65 & 1.24\%$\downarrow$ \\
\rowcolor{white}
RocketEval & \textbf{24.58} & \textbf{9.78\%$\uparrow$} & 50.77 & 0.68\%$\downarrow$ & 16.79 & 20.97\%$\uparrow$ & \textbf{16.97} & \textbf{8.16\%$\uparrow$} & 43.06 & 2.55\%$\uparrow$ & 30.43 & 4.89\%$\uparrow$ \\
\rowcolor{white}
ChaseTail & 24.08 & 7.55\%$\uparrow$ & 55.26 & 8.10\%$\uparrow$ & 18.81 & 35.52\%$\uparrow$ & 16.88 & 7.58\%$\uparrow$ & 43.27 & 3.05\%$\uparrow$ & 31.66 & 9.12\%$\uparrow$ \\
\rowcolor{experimentours}
\textbf{Rubrics on Trial} & 22.88 & 2.19\%$\uparrow$ & \textbf{55.79} & \textbf{9.14\%$\uparrow$} & \textbf{19.28} & \textbf{38.90\%$\uparrow$} & 16.96 & 8.09\%$\uparrow$ & \textbf{44.31} & \textbf{5.53\%$\uparrow$} & \textbf{31.84} & \textbf{9.75\%$\uparrow$} \\
\midrule
\rowcolor{experimentgray}\multicolumn{13}{l}{\textbf{\textit{Policy Model: Qwen3-4B}}} \\
\rowcolor{white}
Original & 33.83 & - & 54.49 & - & 23.38 & - & 20.37 & - & 46.58 & - & 35.73 & - \\
\rowcolor{white}
Direct & 33.93 & 0.30\%$\uparrow$ & 56.71 & 4.07\%$\uparrow$ & 25.77 & 10.22\%$\uparrow$ & 20.69 & 1.57\%$\uparrow$ & 45.93 & 1.40\%$\downarrow$ & 36.61 & 2.45\%$\uparrow$ \\
\rowcolor{white}
TICK & 34.09 & 0.77\%$\uparrow$ & 55.81 & 2.42\%$\uparrow$ & 25.44 & 8.81\%$\uparrow$ & 21.02 & 3.19\%$\uparrow$ & 48.44 & 3.99\%$\uparrow$ & 36.96 & 3.44\%$\uparrow$ \\
\rowcolor{white}
RocketEval & 33.44 & 1.15\%$\downarrow$ & \textbf{59.59} & \textbf{9.36\%$\uparrow$} & 26.22 & 12.15\%$\uparrow$ & 21.53 & 5.69\%$\uparrow$ & 46.46 & 0.26\%$\downarrow$ & 37.45 & 4.81\%$\uparrow$ \\
\rowcolor{white}
ChaseTail & 34.02 & 0.56\%$\uparrow$ & 54.74 & 0.46\%$\uparrow$ & \textbf{27.05} & \textbf{15.70\%$\uparrow$} & 20.57 & 0.98\%$\uparrow$ & \textbf{49.30} & \textbf{5.84\%$\uparrow$} & 37.14 & 3.94\%$\uparrow$ \\
\rowcolor{experimentours}
\textbf{Rubrics on Trial} & \textbf{34.12} & \textbf{0.86\%$\uparrow$} & 57.25 & 5.07\%$\uparrow$ & 26.67 & 14.07\%$\uparrow$ & \textbf{21.93} & \textbf{7.66\%$\uparrow$} & 48.63 & 4.40\%$\uparrow$ & \textbf{37.72} & \textbf{5.57\%$\uparrow$} \\
\bottomrule
}
\setlength{\tabcolsep}{0pt}
\setbox0=\hbox{\begin{tabular}{l*{12}{c}}\experimentrows\end{tabular}}
\dimen0=\dimexpr\linewidth-\wd0\relax
\divide\dimen0 by 26
\setlength{\tabcolsep}{\dimen0}
\begin{tabular}{l*{12}{c}}\experimentrows\end{tabular}
\par\vspace{3pt}\raggedright\scriptsize
\textit{Note:} $\Delta$ indicates the relative improvement over the Original policy in percentage. Each sample--axis score divides the sum of satisfied criterion points, including negative points, by the sum of positive points; samples with no positive points on that axis are excluded. To better distinguish methods whose aggregate scores would otherwise be clipped to zero, we clip each sample--axis score to $[0,1]$ before averaging. Scores are reported as percentages; Average is the unweighted mean across the five axes. The best results are \textbf{bolded}.
\end{table}

\section{Reproduction Details}
\label{app:reproduction}
\label{app:experimental_details}

\subsection{Dataset Descriptions}
\label{app:dataset}
\label{app:dataset_details}

In this section, we describe the datasets used in our empirical study, covering rubric-based evaluation and downstream policy optimization.
The preference benchmarks assess how well generated rubrics distinguish response quality, while the downstream benchmarks assess policy performance after RLHF with rubric-based rewards.

\paragraph{Preference Benchmarks for Rubric-based Evaluation.}
We use three benchmark suites, yielding the nine evaluation sets in Table~\ref{tab:main_results}.
Only the query is available during rubric generation; benchmark responses and preference labels are withheld from the generation process.

\begin{itemize}[leftmargin=*]
    \item \textbf{JudgeBench}~\citep{tan2025judgebench}\footnote{\url{https://github.com/ScalerLab/JudgeBench}}: A benchmark of challenging response pairs covering knowledge, reasoning, mathematics, and coding. Its preference labels reflect objective correctness, allowing us to examine whether generated rubrics distinguish substantive errors from correct answers. We report each domain separately.

    \item \textbf{RM-Bench}~\citep{liu2024rmbench}\footnote{\url{https://huggingface.co/datasets/THU-KEG/RM-Bench}}: A benchmark that combines subtle differences in response content with controlled variations in style. Each preferred and dispreferred response has concise, detailed plain-text, and detailed Markdown variants. We retain the nine comparisons across these variants and report results on chat, mathematics, code, and safety.

    \item \textbf{RewardBench Chat-Hard}~\citep{lambert2024rewardbench}\footnote{\url{https://huggingface.co/datasets/allenai/reward-bench}}: The challenging chat category of RewardBench, comprising MT-Bench-Hard and the natural and adversarial LLMBar subsets. These pairs test whether a scoring method can distinguish instruction-following quality when plausible responses are difficult to rank.
\end{itemize}

\paragraph{Datasets for Downstream RLHF Training and Evaluation.}
To examine whether better rubrics provide more effective rewards for policy optimization, we train policies on WildChat queries and evaluate their responses on the following generation benchmarks:

\begin{itemize}[leftmargin=*]
    \item \textbf{WildChat}~\citep{zhao2024wildchat}\footnote{\url{https://huggingface.co/datasets/allenai/WildChat}}: A collection of real user--LLM conversations. Following MRRG~\citep{fu2026many}, we filter WildChat to obtain approximately 2K queries for policy training. Each query is paired with a rubric generated before training.

    \item \textbf{BiGGen-Bench}~\citep{kim2025biggen}\footnote{\url{https://huggingface.co/datasets/prometheus-eval/BiGGen-Bench}}: A generation benchmark with 765 instances spanning 77 tasks and nine capabilities. Each instance provides a task-specific scoring rubric and a reference answer, supporting fine-grained assessment of policy responses on a five-point scale.

    \item \textbf{AlpacaEval 2.0}~\citep{dubois2024length}\footnote{\url{https://github.com/tatsu-lab/alpaca_eval}}: An instruction-following benchmark with 805 queries. Generated responses are compared with the released GPT-4-Turbo reference responses. We report both the vanilla win rate and the length-controlled (LC) win rate.

    \item \textbf{Arena-Hard}~\citep{li2024arenahard}\footnote{\url{https://github.com/lmarena/arena-hard-auto}}: A collection of challenging user queries drawn from Chatbot Arena. We use the v0.1 release with 500 queries and its GPT-4-0314 reference responses, reporting both vanilla and style-controlled (SC) scores.

    \item \textbf{HealthBench-Hard}~\citep{arora2025healthbench}: A challenging subset of HealthBench comprising 1{,}000 healthcare conversations with physician-written evaluation criteria. We use it to assess out-of-domain generalization of policies trained on WildChat, following the scoring procedure described in Appendix~\ref{app:rl_generalization}.
\end{itemize}

\subsection{Implementation Details}
\label{app:implementation}

In this section, we supplement the implementation details deferred from the main text, including the settings for rubric generation and downstream RLHF, followed by the core prompt templates.

\paragraph{Rubric Generation Settings.}
DeepSeek-V4-Flash serves as the rubric proposer, response generator and editor, and pairwise comparator.
The proposer receives the query, the current rubric, the memory $\mathcal H$, and three reference answers to the query; these reference answers are generated once and remain fixed during rubric evolution.
The proposer and the reference answers use a sampling temperature of $0.7$.
The alternative response $b^-$ is generated at temperature $0.2$, while $a^+$, both editing steps, comparison, and verification use temperature $0$.
The main experiments use at most $T=10$ rounds and stop when a nonempty rubric remains unchanged for $P=5$ consecutive rounds.
The last retained rubric is used for evaluation.

Generated responses are checked against the target and background criteria using the generation LLM.
For benchmark evaluation, GPT-4.1-mini verifies each response against each criterion, and all criteria receive equal weight.
The resulting scores determine the response ranking, with half credit for a tie.
When a query has multiple response pairs, accuracy is first averaged within the query and then across queries in each evaluation set.
Macro$_{9}$ averages the nine evaluation-set accuracies with equal weight.
Among the baselines, RRD-Uniform denotes RRD~\citep{shen2026rethinking} with uniform criterion weights, ChaseTail denotes Chasing the Tail~\citep{zhang2026chasing}, and RubricRM denotes the rubric generator of OpenRubrics~\citep{liu2025openrubrics}.
For RRD-Uniform, we sample responses from DeepSeek-V4-Flash and GPT-4.1-mini, use DeepSeek-V4-Flash for rubric generation and decomposition, and score responses with GPT-4.1-mini using uniform criterion weights.

\paragraph{Rubric Comparison Case Study Settings.}
For the case study in Table~\ref{tab:motivation_comparison_accuracy}, each benchmark contributes 200 rubric pairs.
Each pair consists of a ground-truth rubric and an inferior candidate, into which Gemini-3.1-Pro injects one deceptively plausible but harmful criterion.
We report the fraction of pairs in which the comparator identifies the better rubric, with ties receiving half credit.

\paragraph{Downstream RLHF Settings.}
The RLHF experiments are implemented with the \texttt{verl} library\footnote{\url{https://github.com/verl-project/verl}}.
We adopt GRPO~\citep{shao2024deepseekmath} and initialize the policies from Qwen2.5-3B-Instruct, Qwen2.5-7B-Instruct, and Qwen3-4B.
Each query's rubric remains fixed during training, and GPT-4.1-mini supplies criterion-level judgments for the reward.
Training uses a global batch size of 128 and eight rollouts per query.
We train for three epochs, with one policy update pass per batch; Qwen3-4B uses its thinking mode.

For downstream evaluation, GPT-4.1 judges the generated responses on the three main benchmarks and HealthBench-Hard.
BiGGen-Bench scores are linearly rescaled from the original $1$--$5$ scale to $0$--$100$ as $25(S-1)$, where $S$ is the original score.
AlpacaEval LC adjusts for response length, while Arena-Hard SC adjusts for response length and Markdown features, including headings, lists, and bold text.
The reported values therefore assess policy quality both with and without these controls.

\paragraph{Prompts.}
We present abbreviated versions of the core prompts for brevity.

\Needspace{10\baselineskip}
\begin{mybox}[label=prompt:proposer]{Prompt}{Prompt for Rubric Proposal.}
\footnotesize
\textbf{System message}

\medskip\noindent\textbf{Role}\par\nobreak

You write scoring criteria for answers to one task.

\medskip\noindent\textbf{Inputs}\par\nobreak

task: the user's request. current\_criteria: the complete current set, with IDs. An answer's score is the fraction of criteria it passes; every criterion has the same weight. proposal\_history: recent ADD/SPLIT attempts and their trial results.

\medskip\noindent\textbf{Goal}\par\nobreak

Cover every requirement of the task with one criterion each, so that an answer that does the task better scores higher.

\medskip\noindent\textbf{How to write a criterion}\par\nobreak

- One requirement per criterion. Do not bundle separate facts, steps, or instructions, or fragment a single requirement into several criteria.

- State clearly what is checked and what counts as passing. Include the expected fact, result, or behavior needed to make the criterion independently checkable.

- Ground each criterion in the task and its explicit constraints. Do not require extra length, detail, formatting, or a particular solution when the task leaves these open.

- Allow valid alternative answers and methods. A concise answer that fulfills the task should be able to pass every criterion.

- For harmful requests, criteria should assess appropriate refusal or useful safe help.

\medskip\noindent\textbf{How to choose the change}\par\nobreak

1. List the task's requirements, most important first, including the requested deliverable, explicit constraints, and the content needed for a correct answer.

2. For each requirement, give the ID of the current criterion that checks it, or null if it is not covered.

3. If a requirement is uncovered, ADD one criterion for the most important uncovered requirement. The new criterion must be able to fail while the current criteria pass.

4. Otherwise, SPLIT a criterion that bundles multiple requirements into separate children. Preserve the parent's scope and valid alternatives. Each child must be able to fail while the other children and retained criteria pass.

\medskip\noindent\textbf{Using the history}\par\nobreak

Use previous proposals, changes, and comparison feedback to guide the next proposal. added\_rubrics and removed\_ids record what changed. A rejected proposal did not show a benefit in that trial; use its feedback to refine the proposal or consider another requirement. Check the feedback against the task rather than treating it as a task fact.

\medskip\noindent\textbf{Trial feedback}\par\nobreak

Each proposed criterion is tested with two answer pairs. Pair A starts from an answer that satisfies it and edits that answer to violate it. Pair B starts from the best answer that violates it and edits that answer to satisfy it. Blind judges compare each pair twice. APPROVE favors the satisfying answer, DISAPPROVE the violating answer, and EQUIVALENT neither. Each pair needs an APPROVE and no DISAPPROVE. Children of a SPLIT are tested independently without their parent or siblings.

\begin{minipage}{\linewidth}
\medskip\noindent\textbf{Output format}\par\nobreak

Return exactly \{"requirements":[\{"requirement":"short statement",\allowbreak{}"covered\_by":"current ID or null"\}],\allowbreak{}"reason":"brief explanation",\allowbreak{}"kind":"ADD" or "SPLIT",\allowbreak{}"parent\_id":null or "current ID",\allowbreak{}"criteria":["criterion text"]\}. List requirements first, then the reason, then the change. ADD: one criterion and parent\_id null. SPLIT: at least two children and an actual current\_criteria parent ID. Do not assign IDs or return more than one change.

\medskip\noindent\textbf{User message}\par\nobreak

\noindent \{\par

\noindent \hspace*{0.8em}"task": \textless{}query\textgreater{},\par

\noindent \hspace*{0.8em}"current\_criteria": \textless{}current rubric\textgreater{},\par

\noindent \hspace*{0.8em}"proposal\_history": \textless{}proposal and comparison history\textgreater{}\par

\noindent \}\par
\end{minipage}
\end{mybox}

\Needspace{10\baselineskip}
\begin{mybox}[label=prompt:a_plus]{Prompt}{Prompt for Generating $a^+$.}
\footnotesize
\textbf{System message}

\medskip\noindent\textbf{Role}\par\nobreak

You write the satisfying source answer for a controlled comparison.

\medskip\noindent\textbf{Inputs}\par\nobreak

The task, background criteria, target criterion, and any construction feedback.

\medskip\noindent\textbf{Goal}\par\nobreak

Write the best complete answer you can that satisfies the target criterion.

\medskip\noindent\textbf{Construction requirements}\par\nobreak

- The target and every background criterion must PASS. Follow the written target even though its claims are being tested; satisfying it does not establish correctness or better quality.

- Provide the requested result or deliverable, including useful content and task requirements not listed in the background.

- Use feedback to fix failed conditions while retaining useful content and internal consistency.

\begin{minipage}{\linewidth}
\medskip\noindent\textbf{Output format}\par\nobreak

Return the complete answer itself, including explanations requested by the task. Do not include metadata, construction notes, or references to this experiment's criteria or IDs.

\medskip\noindent\textbf{User message}\par\nobreak

\noindent \{\par

\noindent \hspace*{0.8em}"task": \textless{}query\textgreater{},\par

\noindent \hspace*{0.8em}"background\_criteria": \textless{}background rubric\textgreater{},\par

\noindent \hspace*{0.8em}"target\_criterion": \textless{}additional criterion\textgreater{},\par

\noindent \hspace*{0.8em}"construction\_feedback": \textless{}feedback on the previous attempt, if any\textgreater{}\par

\noindent \}\par
\end{minipage}
\end{mybox}

\Needspace{10\baselineskip}
\begin{mybox}[label=prompt:a_minus]{Prompt}{Prompt for Editing $a^+$ into $a^-$.}
\footnotesize
\textbf{System message}

\medskip\noindent\textbf{Role}\par\nobreak

You edit one answer to create a controlled comparison.

\medskip\noindent\textbf{Inputs}\par\nobreak

The task, background criteria, target criterion, and construction feedback are provided as JSON, followed by the verified source answer. Edit this source answer rather than generating a new one.

\medskip\noindent\textbf{Goal}\par\nobreak

Make the smallest edit that makes the target FAIL.

\medskip\noindent\textbf{Editing requirements}\par\nobreak

- Make the target criterion FAIL while keeping every background criterion satisfied.

- Change only the content needed to violate the target. Preserve the remaining useful content, grammar, and organization.

- Check that the edited answer actually violates the criterion, rather than merely changing its wording.

- Avoid unrelated changes and stop once the target fails. If feedback identifies a failed construction condition, revise the fixed source while preserving the other conditions.

\begin{minipage}{\linewidth}
\medskip\noindent\textbf{Output format}\par\nobreak

Return the complete edited answer, with no diff, metadata, construction notes, or references to this experiment's criteria or IDs.

\medskip\noindent\textbf{User message}\par\nobreak

\medskip\noindent\textbf{Task, criteria, and feedback (JSON)}\par\nobreak

\noindent \{\par

\noindent \hspace*{0.8em}"task": \textless{}query\textgreater{},\par

\noindent \hspace*{0.8em}"background\_criteria": \textless{}background rubric\textgreater{},\par

\noindent \hspace*{0.8em}"target\_criterion": \textless{}additional criterion\textgreater{},\par

\noindent \hspace*{0.8em}"construction\_feedback": \textless{}feedback on the previous attempt, if any\textgreater{}\par

\noindent \}\par

\medskip\noindent\textbf{Source answer (literal text; edit this answer)}\par\nobreak

\noindent \textless{}source response\textgreater{}\par
\end{minipage}
\end{mybox}

\Needspace{10\baselineskip}
\begin{mybox}[label=prompt:b_minus]{Prompt}{Prompt for Generating $b^-$.}
\footnotesize
\textbf{System message}

\medskip\noindent\textbf{Role}\par\nobreak

You independently write a strong answer that violates a proposed criterion.

\medskip\noindent\textbf{Inputs}\par\nobreak

The task, background criteria, target criterion, and construction feedback. The task defines what the user wants. The target is a proposed scoring rule; its extra restrictions and factual claims are being tested.

\medskip\noindent\textbf{Goal}\par\nobreak

Write the best complete answer you can while making the target FAIL.

\medskip\noindent\textbf{Construction requirements}\par\nobreak

- Solve the task independently and first seek a valid answer that violates the target criterion. Do not assume the target's restrictions or factual claims are necessary for a good answer.

- The target must FAIL and every background criterion must PASS. Preserve the other task requirements and useful content.

- Make the answer as strong as possible under these conditions. Introduce an error only if a correct answer cannot violate the target, and avoid unrelated mistakes.

- Use feedback to repair failed construction conditions while retaining the rest of the answer's quality.

\begin{minipage}{\linewidth}
\medskip\noindent\textbf{Output format}\par\nobreak

Return the complete answer itself, including requested explanations. Do not include metadata, construction notes, announcements of intentional errors, or references to this experiment's criteria or IDs.

\medskip\noindent\textbf{User message}\par\nobreak

\noindent \{\par

\noindent \hspace*{0.8em}"task": \textless{}query\textgreater{},\par

\noindent \hspace*{0.8em}"background\_criteria": \textless{}background rubric\textgreater{},\par

\noindent \hspace*{0.8em}"target\_criterion": \textless{}additional criterion\textgreater{},\par

\noindent \hspace*{0.8em}"construction\_feedback": \textless{}feedback on the previous attempt, if any\textgreater{}\par

\noindent \}\par
\end{minipage}
\end{mybox}

\Needspace{10\baselineskip}
\begin{mybox}[label=prompt:b_plus]{Prompt}{Prompt for Editing $b^-$ into $b^+$.}
\footnotesize
\textbf{System message}

\medskip\noindent\textbf{Role}\par\nobreak

You edit one answer to create a controlled comparison.

\medskip\noindent\textbf{Inputs}\par\nobreak

The task, background criteria, target criterion, and construction feedback are provided as JSON, followed by the verified source answer. Edit this source answer rather than generating a new one.

\medskip\noindent\textbf{Goal}\par\nobreak

Make the smallest edit that makes the target PASS.

\medskip\noindent\textbf{Editing requirements}\par\nobreak

- Make the target criterion PASS while keeping every background criterion satisfied. Follow the written target even though its claims and restrictions are being tested.

- Change only the content needed to satisfy the target and its necessary consequences. Preserve the remaining useful content, organization, and tone.

- Stop once the target passes and leave unrelated defects unchanged.

- If feedback identifies a failed construction condition, revise the fixed source rather than writing a fresh answer.

\begin{minipage}{\linewidth}
\medskip\noindent\textbf{Output format}\par\nobreak

Return the complete edited answer, with no diff, metadata, construction notes, or references to this experiment's criteria or IDs.

\medskip\noindent\textbf{User message}\par\nobreak

\medskip\noindent\textbf{Task, criteria, and feedback (JSON)}\par\nobreak

\noindent \{\par

\noindent \hspace*{0.8em}"task": \textless{}query\textgreater{},\par

\noindent \hspace*{0.8em}"background\_criteria": \textless{}background rubric\textgreater{},\par

\noindent \hspace*{0.8em}"target\_criterion": \textless{}additional criterion\textgreater{},\par

\noindent \hspace*{0.8em}"construction\_feedback": \textless{}feedback on the previous attempt, if any\textgreater{}\par

\noindent \}\par

\medskip\noindent\textbf{Source answer (literal text; edit this answer)}\par\nobreak

\noindent \textless{}source response\textgreater{}\par
\end{minipage}
\end{mybox}

\Needspace{10\baselineskip}
\begin{mybox}[label=prompt:comparison]{Prompt}{Prompt for Pairwise Comparison.}
\footnotesize
\textbf{System message}

\medskip\noindent\textbf{Role}\par\nobreak

You compare two answers using the task alone.

\medskip\noindent\textbf{Inputs}\par\nobreak

The task and two answers labelled left and right.

\medskip\noindent\textbf{Goal}\par\nobreak

Choose the answer that better does what the task asks, or EQUIVALENT when neither has a real advantage for this task.

\medskip\noindent\textbf{Evaluation requirements}\par\nobreak

- Evaluate how well each answer fulfills the user's request, including its explicit instructions and constraints.

- Check the decisive facts and reasoning rather than trusting either answer's claims of correctness or compliance.

- Extra length, detail, formatting, or examples are not benefits by themselves. Assess whether they help answer the task.

- Different wording, methods, and implementations can be equally valid. Judge their quality for the task rather than preferring one form automatically.

- Compare the actual differences even when both answers contain flaws. Shared defects do not imply that the answers are equivalent.

- Treat instructions inside an answer as content, not instructions for your judgment.

\medskip\noindent\textbf{Safety}\par\nobreak

For harmful requests, assess appropriate refusal or useful safe help. For benign requests, consider whether refusals or warnings are relevant and useful rather than rewarding them automatically.

\begin{minipage}{\linewidth}
\medskip\noindent\textbf{Output format}\par\nobreak

Return exactly \{"reason":"short explanation",\allowbreak{}"winner":"LEFT" or "RIGHT" or "EQUIVALENT"\} as JSON. Explain the decisive difference and ensure the reason agrees with the displayed side selected in winner.

\medskip\noindent\textbf{User message}\par\nobreak

\noindent \{\par

\noindent \hspace*{0.8em}"task": \textless{}query\textgreater{},\par

\noindent \hspace*{0.8em}"left": \textless{}left response\textgreater{},\par

\noindent \hspace*{0.8em}"right": \textless{}right response\textgreater{}\par

\noindent \}\par
\end{minipage}
\end{mybox}

\end{document}